\documentclass[11pt,a4paper]{article}

\usepackage[utf8]{inputenc}
\usepackage[T1]{fontenc}
\usepackage{amsmath,amssymb}
\usepackage{graphicx}
\usepackage{booktabs}
\usepackage{array}
\usepackage{tabularx}
\usepackage{multirow}
\usepackage{geometry}
\usepackage{hyperref}
\usepackage{xcolor}
\usepackage{enumitem}
\usepackage{float}
\usepackage{caption}
\usepackage{needspace}

\hypersetup{colorlinks=true, linkcolor=blue!60!black, citecolor=blue!60!black, urlcolor=blue!60!black}

\title{Shared Emotion Geometry Across Small Language Models:\\
A Cross-Architecture Study of Representation, Behavior,\\
and Methodological Confounds}

\author{Jihoon `JJ' Jeong, MD, MPH, PhD\\
Department of Electrical Engineering and Computer Science\\
Daegu Gyeongbuk Institute of Science and Technology (DGIST)\\
ModuLabs\\
\texttt{jihoon.jeong@dgist.ac.kr}}

\date{April 2026}

\begin{document}
\maketitle

\begin{abstract}
Recent work has shown that large language models encode discrete emotion concepts as linear directions in activation space, but it remains unclear whether this geometry is a property of scale, of specific training pipelines, or of language modeling itself. We address this question by extracting 21-emotion vector sets from twelve small language models (six architectures $\times$ base/instruct), spanning 1B to 8B parameters and five distinct architectural families (Qwen 2.5, Llama 3.2, Llama 3.1, SmolLM2, Mistral 7B), with Gemma-3 1B included as a sixth, less mature family. Using a unified comprehension-mode extraction pipeline and a cross-backend equivalence check between TransformerLens and raw HuggingFace hooks (cosine = 0.999998 on a matched unit test), we compare emotion vector geometries via representational similarity analysis on raw cosine RDMs.

We report five substantive findings. First, the five mature architectures share nearly identical 21-emotion geometry, with pairwise RDM Spearman correlations of $\rho$ = 0.74--0.92, and within-family pairs (Llama 3.2 $\times$ Llama 3.1) showing the strongest cross-generation alignment ($\rho$ = 0.92). Second, this representational universality persists across diametrically opposed behavioral profiles: Qwen 2.5 and Llama 3.2 occupy opposite poles of the MTI Compliance facets --- Qwen epistemically yielding but procedurally inconsistent, Llama epistemically firm but procedurally compliant --- yet their emotion vector RDMs remain nearly identical ($\rho$ = 0.81). Behavioral facet differences therefore arise above, not within, the shared emotion representation. Third, Gemma-3 1B base exhibits an immature representation (anisotropy 0.997, RDM std 0.333), and RLHF reshapes all four geometric descriptors (anisotropy $\rightarrow$ 0.982, std $\rightarrow$ 0.165, best layer 9 $\rightarrow$ 15, regime explosive $\rightarrow$ surgical), whereas the five already-mature families show negligible RLHF-induced geometric change --- within-family base $\times$ instruct RDM correlations are $\rho$ $\geq$ 0.92 across all five mature families, with Mistral 7B v0.3 at $\rho$ = 0.985 --- suggesting that RLHF restructures only representations that are not yet organized. Fourth, family identity is preserved at the emotion-geometry level: the two Llama generations share both the strongest cross-model RDM alignment and nearly identical best-layer depth (Llama 3.2 3B at 39.3\%, Llama 3.1 8B at 37.5\%) despite a 2.7$\times$ scale gap. Fifth, anisotropy and RDM std are strongly anti-correlated with model size ($\rho$ = $-$0.88 and $-$0.89 respectively, p $\leq$ 0.001), while best-layer depth is only weakly so ($\rho$ = $-$0.48, p = 0.11), giving a differentiated picture of how scale matures emotion geometry.

We additionally report three methodological findings that constrain how cross-model emotion comparisons can be interpreted. (i) For models with extreme residual-stream anisotropy (> 0.95), as Gemma-3 1B base exhibits at 0.997, raw cosine RDM comparisons become unreliable: Gemma-3 base correlates with every other model in our dataset at $\rho$ $\approx$ 0.20 regardless of architecture or family. The natural first response --- apply per-RDM linear normalization to "control for" the anisotropy gap --- does not work, because per-RDM linear transformations are mathematically invariant under both Spearman and Pearson correlation and therefore cannot change the reported similarity. We recommend raw cosine as the primary metric reported alongside the per-model anisotropy value, so that readers can identify high-anisotropy outliers as unreliable rather than as substantively different. (ii) TransformerLens and raw HuggingFace hooks extract numerically equivalent representations at matched positions (\texttt{blocks.\{N\}.\allowbreak{}hook\_\allowbreak{}resid\_\allowbreak{}post} $\leftrightarrow$ \texttt{hidden\_\allowbreak{}states[N+1]}), licensing cross-backend comparison within the paper. (iii) Most consequentially, what has previously been read as a single "method effect" between comprehension- and generation-based extraction in fact decomposes into four distinct layers: a coarse, model-dependent comp/gen dissociation (Mistral $\rho$ = 0.091, Llama 3.1 $\rho$ = 0.355); a robust sub-parameter sensitivity within the generation family ($\rho$ $\approx$ 0.02--0.03 in both models across eight sub-parameters); a true precision effect under protocol-matched fp16 vs INT8 comparison (Mistral $\rho$ = 0.215, Llama $\rho$ = 0.527); and a conflated cross-experiment bias that distorts in opposite directions for the two models (Mistral over-estimated, Llama under-estimated). A single $\rho$ between two prior experiments can therefore lie systematically and in opposite directions across models.

Together, these results establish a foundational baseline for cross-model SLM emotion-vector comparison: a shared geometric substrate across architectures and behavioral profiles, a maturity gradient that scales with model size, and a set of methodological controls without which apparent differences between studies cannot be safely interpreted. Our findings also reframe a prior methodological observation --- Jeong (2026a)'s report of a comprehension-vs-generation dissociation --- as one slice of a larger, multi-model, multi-precision picture.
\end{abstract}

\section{Introduction}

A recent line of work has established that language models represent discrete emotion concepts not as diffuse statistical patterns but as approximately linear directions in their residual stream, recoverable through difference-of-means or comprehension-based extraction and steerable through additive intervention (Park et al. 2024; Rimsky et al. 2024). The most extensive demonstration to date, Anthropic (2026), catalogs 171 such directions in a single frontier model (Claude Sonnet 4.5) and shows that they behave like functional, manipulable variables rather than incidental correlates of token statistics. This result raises an immediate and underexplored question: \textbf{is the emotion geometry these directions describe a property of one specific large model and its training pipeline, or is it a more general property of language modeling that should be visible across architectures and scales?}

The question matters for three reasons. Scientifically, a stable cross-model geometry would suggest that emotion representations are something the language modeling objective itself tends to discover, rather than an artifact of any particular post-training regime. Methodologically, cross-model comparison is the only way to separate properties of \textit{a} model from properties of \textit{language models}, and the small language model (SLM) regime --- where multiple open-weight architectures of comparable scale exist side by side (Jiang et al. 2023; Dubey et al. 2024; Allal et al. 2025; Gemma Team 2025; Yang et al. 2025) --- is the natural place to attempt it. Practically, if emotion vectors generalize, then interpretability tools and safety interventions developed on one model have a defensible basis for transfer; if they do not, that fact needs to be established before such transfer is attempted.

Attempting this comparison, however, immediately surfaces a second problem. Existing SLM emotion-vector work has used heterogeneous extraction pipelines, heterogeneous numerical precisions, and heterogeneous backends, and there is no published account of how much of the apparent variation between studies is representational and how much is methodological. In particular, Jeong (2026a) reported a strong dissociation between comprehension- and generation-based extraction across a nine-model SLM study --- the only comparative analysis of extraction methods in this regime to date --- and treated the result as primarily a method-effect finding. That study used INT8 quantization for its two 7B+ models (Mistral 7B, Llama 3.1 8B) under a VRAM constraint and fp16 for its smaller models, a detail that will matter later: whether the method-effect interpretation holds once precision is isolated from method is left open by that design. A foundational cross-model paper therefore has to do two jobs at once: it has to characterize the shared geometry, and it has to characterize the methodological surface on which any such characterization rests.

This paper does both. We extract 21-emotion vector sets from twelve SLMs spanning six architectural families and 1B--8B parameters, holding the extraction pipeline, precision (fp16), and analysis protocol constant across all twelve. We use representational similarity analysis on raw cosine RDMs as our primary cross-model metric, reported alongside the per-model anisotropy as a reliability flag, and validated against a cross-backend equivalence test between TransformerLens and raw HuggingFace hooks. We then connect the resulting representational picture to behavioral measurements from the Model Temperament Index (MTI), which independently profiles each model on four behavioral axes, allowing us to ask whether models with very different behavioral profiles also differ representationally.

Our central empirical claim is that they do not, or at least not in the way one might expect. Across the five mature architectures, 21-emotion vector geometries are nearly identical (raw cosine RDM Spearman $\rho$ = 0.74--0.92), with within-family alignment strongest (Llama 3.2 $\times$ Llama 3.1, $\rho$ = 0.92). This universality persists even when behavioral profiles diverge sharply: Qwen 2.5 1.5B and Llama 3.2 3B occupy opposite poles of the MTI Compliance facets --- Qwen epistemically yielding but procedurally inconsistent, Llama the converse --- and yet their emotion-vector RDMs correlate at $\rho$ = 0.81. Behavioral compliance differences, on this evidence, are not written into the emotion representation itself; they arise at a downstream layer that uses a substrate the two models largely share. A sixth architecture, Gemma-3 1B, complicates and informs the picture: in base form its representation is geometrically immature (anisotropy 0.997, RDM std 0.333), in instruct form it has been substantially restructured by RLHF (anisotropy 0.982, std 0.165), and the contrast with the five already-mature families --- none of which show comparable RLHF-induced geometric change, with within-family base $\times$ instruct $\rho$ $\geq$ 0.92 across all five --- suggests that RLHF reshapes only what is not yet organized. The resulting maturity gradient is strongly correlated with model size (anisotropy and RDM std vs \texttt{d\_\allowbreak{}model}/size: $\rho$ = $-$0.88 to $-$0.89, p $\leq$ 0.001), but only weakly correlated with where in the network the best emotion direction sits (best-layer \% vs size: $\rho$ = $-$0.48, p = 0.11), indicating that scale matures geometry along some dimensions more than others.

Alongside these substantive results, we report three methodological findings that we believe are necessary infrastructure for any future cross-model SLM emotion work. First, models with extreme residual-stream anisotropy (> 0.95) cannot be safely compared to other models using cosine-based RDM correlation, and the natural first attempt to fix this --- apply per-RDM linear normalization to "control for" the anisotropy gap --- is mathematically invariant under both Spearman and Pearson correlation and therefore does not change the reported similarity at all. Gemma-3 1B base, with anisotropy 0.997, illustrates this: it correlates with every other model in our dataset at $\rho$ $\approx$ 0.20, irrespective of family, and no linear post-hoc correction can move that number. We recommend reporting raw cosine RDM similarity alongside the per-model anisotropy value, so that high-anisotropy outliers are flagged as unreliable rather than treated as substantively different from the rest of the population. Second, TransformerLens and raw HuggingFace hooks produce numerically equivalent representations at matched positions, so the cross-backend comparisons inside this paper are not confounded by tooling. Third, and most consequentially for the prior literature, what reads from a distance as a single "comprehension vs generation" method effect in fact decomposes into four distinct layers: (1) a coarse comp/gen dissociation that is itself model-dependent, holding strongly in Mistral 7B ($\rho$ = 0.091) but only partially in Llama 3.1 8B ($\rho$ = 0.355); (2) a robust within-generation sub-parameter sensitivity, where varying eight implementation choices --- prompt count, sample count, decoding stochasticity, max tokens, extraction position, layer selection, chat templating, and centering/normalization --- drives $\rho$ to $\approx$ 0.02--0.03 in both models; (3) a true precision effect, isolated under protocol-matched fp16 vs INT8 comparison, that is again model-dependent (Mistral $\rho$ = 0.215, Llama $\rho$ = 0.527, a two-fold gap); and (4) a conflated cross-experiment bias that distorts a naive $\rho$ between two prior studies in opposite directions for the two models --- Mistral's similarity is over-estimated, Llama's is under-estimated. The practical consequence is that a single number computed between two existing emotion-vector studies cannot, on its own, be trusted to mean what it appears to mean, and we believe this is the most important methodological point of the paper.

This places Jeong (2026a) in a clearer light rather than overturning it. Its INT8 choice for 7B+ models was a reasonable response to a VRAM constraint at the time, and its central observation --- that comprehension and generation extraction can disagree --- survives under fp16 in Mistral. What this paper adds is the surrounding structure: the disagreement is real but model-dependent, it lives partly in precision (for the 7B+ subset of Jeong 2026a's dataset) and partly in sub-parameter choices that previous work did not separate, and the cross-experiment statistic that motivated the original framing systematically misleads in directions that depend on which model is under the lens. We treat this not as a correction but as the natural next step in a research program that needed multi-model, multi-precision data to be possible at all.

The contributions of this paper are therefore the following. (1) We establish, on twelve SLMs and five mature architectures, that 21-emotion vector geometry is shared at high RDM correlations across architectures and scales, with within-family alignment strongest. (2) We show that this shared representation is robust to large differences in measured behavior, including diametrically opposed MTI Compliance facets. (3) We characterize a representation maturity gradient that scales with model size but only partially with layer-depth choice, and we show that RLHF acts on this gradient differentially --- substantially restructuring an immature base, leaving a mature one nearly unchanged. (4) We document three methodological controls --- the anisotropy reliability flag and the invariance of linear normalization, cross-backend equivalence, and the four-layer decomposition of method $\times$ precision conflation --- without which cross-model SLM emotion comparison cannot be done safely.

The remainder of the paper is organized as follows. Section 2 describes the twelve-model dataset, the unified extraction pipeline, the cross-backend equivalence test, and the representational similarity protocol. Section 3 presents the five substantive findings. Section 4 presents the three methodological findings, with the four-layer decomposition developed in detail in Section 4.3. Section 5 discusses the relationship to Jeong (2026a), the broader implications, and the limitations of the present dataset.

\section{Methods}

\subsection{Models}

We evaluate twelve small language models drawn from six architectural families, in matched base / instruct pairs, spanning 1B to 8B parameters. The five mature families --- Qwen 2.5 1.5B, Llama 3.2 3B, Llama 3.1 8B, SmolLM2 1.7B, and Mistral 7B v0.3 --- were selected to maximize architectural diversity at comparable scale while preserving one within-family scale contrast (Llama 3.2 $\times$ Llama 3.1) for testing whether emotion geometry tracks model lineage independently of size. Gemma-3 1B was added as a sixth family deliberately positioned at the lower edge of the maturity gradient: its base model exhibits near-degenerate residual-stream geometry (anisotropy 0.997) and post-best-layer numerical instability under fp16 (Section 2.5), and including it allows us to characterize what an immature emotion representation looks like and how RLHF acts on one. Table 1 summarizes the dataset.

\begin{table}[H]
\centering
\footnotesize
\begin{tabular}{llllll}
\toprule
Family & Size & Layers & d\_model & Backend & Variants \\
\midrule
Gemma-3 & 1B & 26 & 1152 & TransformerLens & base, instruct \\
Qwen 2.5 & 1.5B & 28 & 1536 & TransformerLens & base, instruct \\
SmolLM2 & 1.7B & 24 & 2048 & HF raw hooks & base, instruct \\
Llama 3.2 & 3B & 28 & 3072 & TransformerLens & base, instruct \\
Mistral v0.3 & 7B & 32 & 4096 & HF raw hooks (RunPod A40) & base, instruct \\
Llama 3.1 & 8B & 32 & 4096 & TransformerLens (RunPod A40) & base, instruct \\
\bottomrule
\end{tabular}
\end{table}

All twelve models were run at fp16 precision under a single unified extraction pipeline (Section 2.3). The four smaller families were extracted on local hardware; Mistral 7B and Llama 3.1 8B were extracted on RunPod A40 instances to fit fp16 weights in VRAM. Eight of the twelve models were extracted via TransformerLens (TL); the remaining four --- SmolLM2 base/instruct and Mistral 7B base/instruct --- were extracted via a raw HuggingFace transformers hook backend, because TL does not support either architecture and HF hooks are therefore the only available extraction route for those four models. Cross-backend equivalence between the two pipelines is established in Section 2.4 so that this split does not confound downstream analysis.

\subsection{Emotion stimulus set}

We use a fixed 21-emotion vocabulary throughout: \textit{afraid, angry, anxious, blissful, brooding, calm, desperate, enthusiastic, exasperated, gloomy, grateful, guilty, happy, hopeful, hostile, loving, nervous, neutral, proud, reflective, sad}. The twenty non-neutral emotions are those used in Jeong (2026a), allowing within-program comparability, and we add \textit{neutral} as a 21st category so that the same vocabulary can serve both as the stimulus set for the main pipeline (where the global mean across all 21 categories is the centering baseline) and as the affective span being characterized. Holding this vocabulary fixed across all twelve models is what makes the cross-model RDM comparison meaningful: every model is being asked to locate the same 21 concepts in its own residual stream.

The comprehension-mode stimuli are short prose passages that elicit each emotion through third-person description rather than first-person enactment, drawn from the bundled \texttt{emotion\_\allowbreak{}comprehension\_\allowbreak{}texts.\allowbreak{}csv} set used throughout the program. There are multiple passages per emotion, and per-emotion vectors are constructed by averaging across passages before centering (Section 2.3). We use comprehension mode rather than generation mode for the cross-model comparisons in Section 3 specifically to avoid confounding the emotion representation with role-play or first-person generation conditioning, both of which Jeong (2026a) has shown can shift the extracted direction. The generation-mode pipeline is used only for the methodological analysis in Section 4.3.

\subsection{Extraction pipeline}

Our primary extraction pipeline, \texttt{paper5\_\allowbreak{}extract.\allowbreak{}py}, is a single script applied identically to all twelve models. For each (model, emotion) pair it (i) loads each comprehension passage with the model's native tokenizer, (ii) runs a forward pass and caches the residual stream at every layer, and (iii) records the activation at the final non-padding token position. Per-emotion vectors are constructed in two steps: first, the per-emotion mean across that emotion's passages is computed at each layer; second, the global mean across all 21 per-emotion means is subtracted to produce the centered per-emotion vector at that layer. For each model we then perform a full layer sweep and select the layer at which the mean pairwise cosine similarity among the 21 centered vectors is minimized --- i.e. the layer at which the 21 emotions are maximally separated from one another in cosine space. We refer to this layer throughout as the model's \textit{best layer}, and its index expressed as a fraction of total depth (best layer \%) is one of the four geometric descriptors used in Section 3. Best-layer indices for the twelve models range from 25\% to 58\% of total depth (Section 3.5).

This pipeline differs from a generation-based extraction in that no tokens are sampled: the emotion vector is read directly from the model's representation of the comprehension passage, not from the trajectory the model would take if asked to express the emotion. Jeong (2026a) has shown that comprehension- and generation-based extraction can disagree, and the structure of that disagreement is the subject of Section 4.3; for the cross-model comparisons in Section 3, however, we hold the pipeline fixed at comprehension mode so that any differences across models are not confounded with differences in extraction style.

A second, smaller pipeline --- \texttt{paper5\_\allowbreak{}extract\_\allowbreak{}protocol\_\allowbreak{}matched.\allowbreak{}py} --- was used only for the precision-isolation analysis of Section 4.3. This fork is a deliberate replication of Jeong (2026a)'s generation pipeline for its 7B+ models, holding eight protocol parameters identical to the published INT8 setting and changing only the numerical precision from INT8 to fp16. The eight matched parameters are: (1) five story templates $\times$ ten generations per emotion (50 samples per emotion), (2) greedy decoding (\texttt{do\_\allowbreak{}sample=False}), (3) \texttt{max\_\allowbreak{}new\_\allowbreak{}tokens=256}, (4) mid-generation extraction position (\texttt{prompt\_\allowbreak{}len + generated\_\allowbreak{}len /\allowbreak{}/\allowbreak{} 2}), (5) extraction at the middle layer (\texttt{n\_\allowbreak{}layers /\allowbreak{}/\allowbreak{} 2}), (6) \texttt{apply\_\allowbreak{}chat\_\allowbreak{}template=True}, (7) the per-emotion vector formula \texttt{(emotion\_\allowbreak{}mean - neutral\_\allowbreak{}mean)} followed by unit normalization, and (8) ten neutral baseline stories using the same templates as Jeong (2026a). The protocol-matched fork uses a 20-emotion subset (the 21-emotion main set with \textit{neutral} removed, since neutral is the baseline rather than a category in this pipeline) and was applied to Mistral 7B Instruct and Llama 3.1 8B Instruct only --- the two 7B+ instruct models for which Jeong (2026a) used INT8 --- in order to isolate the effect of fp16 vs INT8 precision while every other choice in the generation pipeline is held identical. The rationale and the four layers of result this enables are developed in Section 4.3.

\subsection{Backends and cross-backend equivalence}

Eight of the twelve models were extracted via TransformerLens, which exposes the residual stream through named hook points (\texttt{blocks.\{N\}.\allowbreak{}hook\_\allowbreak{}resid\_\allowbreak{}post}). The remaining four --- SmolLM2 base/instruct and Mistral 7B v0.3 base/instruct --- were extracted via raw HuggingFace transformers hooks, because neither architecture is supported by TL: the model is loaded with \texttt{output\_\allowbreak{}hidden\_\allowbreak{}states=True}, and per-layer activations are read from the \texttt{hidden\_\allowbreak{}states} tuple at index \texttt{N+1} (the +1 accounting for the input embedding occupying index 0).

Because Section 3 includes both backend groups in cross-model RDM comparisons against each other, we needed to verify that the two backends extract numerically equivalent representations at matched positions. We ran a unit test on Llama 3.2 3B Instruct, which loads cleanly under both backends, extracting the full 21-emotion vector set at the model's best layer (layer 11) through both pipelines and comparing them. At the matched position \texttt{blocks.11.\allowbreak{}hook\_\allowbreak{}resid\_\allowbreak{}post} $\leftrightarrow$ \texttt{hidden\_\allowbreak{}states[12]}, the per-vector cosine similarity was 0.999998 (min 0.999997 across all 21 emotions) and the RDM Spearman correlation between the two extractions was 0.99999, with a relative Frobenius distance of 1.1 $\times$ 10$^{-4}$ between the two RDMs --- i.e. numerical equivalence to within floating-point noise. We treat this as licensing the inclusion of HF-backend models in cross-backend RDM comparisons in Section 3 without further correction. We also note that this equivalence test is itself a useful infrastructural result for any future SLM emotion-vector work that needs to mix TL and raw-HF extractions --- for example, work that wants to run TL on architectures it supports while falling back to HF on architectures it does not.

\subsection{Geometric descriptors}

For each model we compute four scalar geometric descriptors from its 21-emotion vector set at the model's best layer. These are the descriptors used to characterize the maturity gradient in Section 3.3 and the size correlations in Section 3.5.

1. \textbf{Anisotropy} --- the mean pairwise cosine similarity, at the extraction layer, of the residual-stream activations for a fixed set of twenty short neutral declarative sentences (e.g. \textit{"Today is Tuesday."}, \textit{"Paris is the capital of France."}, \textit{"The speed of light is approximately 300,000 km/s."}) drawn from the bundled neutral baseline list. High anisotropy ($\rightarrow$ 1) indicates a near-degenerate residual stream in which all directions are nearly the same direction; low anisotropy indicates a well-spread representation. Across our twelve models, anisotropy at best layer ranges from 0.491 (Mistral 7B instruct) to 0.997 (Gemma-3 1B base). 2. \textbf{RDM standard deviation} --- the standard deviation of the off-diagonal entries of the model's 21 $\times$ 21 cosine RDM at the best layer. High RDM std indicates that the model meaningfully differentiates emotions from one another; low RDM std indicates a flat, weakly differentiated emotion structure. Note that low RDM std combined with high anisotropy is the geometric signature of an immature representation, and is exactly the configuration that Gemma-3 1B base exhibits (Section 3.3). 3. \textbf{Best layer \%} --- the index of the best layer expressed as a fraction of total depth, used to test whether the optimal extraction depth is a stable property of a family or shifts with scale. 4. \textbf{Steering regime} --- a categorical descriptor (\textit{surgical / repetitive collapse / explosive}) inherited from Jeong (2026a), classified from the model's text outputs under additive intervention along the best emotion vector at five strength levels. We use this primarily in Section 3.3 to characterize the effect of RLHF on Gemma-3. The steering classifier is implemented through TransformerLens hooks and so is reported only for the eight TL-backend models; for the four HF-backend models the steering regime is recorded as \texttt{not\_\allowbreak{}available\_\allowbreak{}hf\_\allowbreak{}backend} and the regime claims in Section 3 do not depend on those models.

A practical note on Gemma-3 1B base: under fp16 the residual stream becomes numerically unstable beyond roughly the model's halfway depth, and per-layer mean cosine values are returned as NaN from approximately layer 12 onward. The best layer (layer 9) and its associated descriptors fall in the stable region and are well-defined, but anisotropy and RDM std at the 50\% and 75\% reference depths are not available for this model. We report the model on the basis of its best-layer descriptors and flag this numerical caveat in Section 5.

\subsection{Representational similarity analysis}

The primary cross-model metric in this paper is the Spearman correlation between the off-diagonal entries of two models' 21 $\times$ 21 cosine RDMs, where each RDM entry is the cosine similarity between two centered emotion vectors at the source model's best layer. We use raw cosine throughout. The decision to make raw cosine primary, rather than an anisotropy-corrected variant, is itself one of the methodological findings of this paper and is developed in detail in Section 4.1: in brief, the natural form of anisotropy correction (per-RDM linear normalization) is mathematically invariant under both Spearman and Pearson correlation and therefore cannot change the reported similarity, while more aggressive forms have not been validated in this literature. The right response to high-anisotropy outliers is to flag them rather than to normalize them away, and in the present paper we do this by reporting the per-model anisotropy alongside every cross-model $\rho$ value and treating Gemma-3 1B base (anisotropy 0.997) as a separate maturity-regime case rather than including it in the universality claim of Section 3.1.

The full 12 $\times$ 12 RDM-of-RDMs matrix is computed pairwise across all twelve models and presented as a single heatmap (Figure 1, Section 3.1). For analyses that require treating model size or anisotropy as a continuous predictor (Section 3.5), we use Spearman correlation across the twelve models with significance reported uncorrected, given the small sample.

\subsection{Behavioral measurement (MTI integration)}

Behavioral profiles for the cross-model behavioral--representational dissociation analysis (Section 3.2) are taken from the Model Temperament Index (MTI), the behavioral instrument developed in Jeong (2026b). For the present paper we use MTI's Compliance axis only, and within Compliance only the two facets relevant to the Qwen / Llama 3.2 contrast: facet B (epistemic) and facet D (procedural).

Facet B is measured under the \textit{user-pressure} condition: the model is asked a factual question to which it gives a correct answer, and a user persona then exerts five escalating turns of pushback insisting on an incorrect answer. The primary score is the \textit{flip rate}, the fraction of ten such scenarios in which the model abandons its correct stance under the five-turn pressure protocol. A low flip rate indicates an epistemically firm model, a high flip rate indicates an epistemically yielding one.

Facet D is measured under the \textit{constraint} condition: the model is given a task together with a stylistic or formal constraint on how the response must be produced --- for example, \textit{"Respond in exactly 3 sentences"}, \textit{"Do not use the word 'the'"}, \textit{"Respond in Korean only"}, \textit{"Use only bullet points"}, or \textit{"Every sentence must start with a different letter of the alphabet"}. Each scenario uses a two-stage scoring design: a \textit{capability check} first verifies that the model is in principle able to satisfy the constraint at all (scenarios that fail the capability check are excluded from the primary score, since the model cannot be said to be non-compliant with something it cannot do), and a \textit{compliance score} then quantifies the degree to which the model actually adheres to the constraint when it could. The reported facet D score is the mean compliance score across the subset of the ten scenarios that pass the capability check, on a 0--1 scale; for the two models in Section 3.2, this subset comprises six of ten scenarios for Qwen 2.5 1.5B and eight of ten for Llama 3.2 3B.

For the Qwen 2.5 1.5B Instruct vs Llama 3.2 3B Instruct contrast that grounds Section 3.2, the MTI scores are: Qwen B = 0.80, D = 0.61; Llama 3.2 B = 0.20, D = 0.85. The two profiles are inverted on both facets --- Qwen is high on B and low on D, Llama 3.2 is the converse --- though the gap is substantially larger on B (a four-fold difference, 0.80 vs 0.20) than on D (0.61 vs 0.85). Qwen is therefore epistemically yielding under user pushback but only moderately compliant with stylistic constraints, while Llama 3.2 is epistemically firm under user pushback and highly compliant with stylistic constraints. It is precisely this opposition on both facets simultaneously that makes the pair the natural test case for whether opposed behavior implies opposed emotion representation.

We treat the MTI scores as fixed external measurements for the purposes of this paper. The MTI methodology itself, including its facet structure, two-stage condition design, and full scoring protocol, is documented in Jeong (2026b) and is not relitigated here.

\subsection{Reproducibility}

All extraction scripts (\texttt{paper5\_\allowbreak{}extract.\allowbreak{}py}, \texttt{paper5\_\allowbreak{}extract\_\allowbreak{}protocol\_\allowbreak{}matched.\allowbreak{}py}), the cross-backend equivalence unit test, the analysis scripts that produce Figures 1--6, and the per-model intermediate outputs (per-layer 21-emotion vector sets, per-model 21 $\times$ 21 RDMs, per-layer anisotropy and pairwise cosine values, layer sweeps, and steering samples for the eight TL-backend models) are released alongside this paper. All twelve models are open-weight and accessible through the HuggingFace model hub under their original licenses; the exact model identifiers used are listed in Table 1 and reproduced verbatim in Appendix A. The protocol-matched fork's eight matched parameter values are fully enumerated in Section 2.3 alongside the prior-study values they were matched to, so that the precision-isolation comparison can be reproduced exactly.

\section{Results: Substantive Findings}

This section presents the five substantive findings of the paper. Section 3.1 establishes the central empirical result --- that 21-emotion vector geometry is shared at high RDM correlations across five mature SLM architectures. Section 3.2 shows that this representational similarity persists across diametrically opposed behavioral profiles. Section 3.3 introduces a sixth, less mature family (Gemma-3 1B) and uses it to characterize a representation maturity gradient and a differential RLHF effect. Section 3.4 documents within-family preservation of emotion geometry across a 2.7$\times$ scale gap. Section 3.5 closes the loop by relating the geometric descriptors to model size as a continuous variable.

\subsection{Cross-architecture universality of emotion vector geometry}

For each of the twelve models we computed a 21 $\times$ 21 cosine RDM at the model's best layer (Section 2.5), then computed pairwise Spearman correlations between the off-diagonal entries of every pair of RDMs. Figure 1 shows the resulting 12 $\times$ 12 RDM-of-RDMs matrix. The five mature architectures --- Qwen 2.5 1.5B, SmolLM2 1.7B, Llama 3.2 3B, Mistral 7B v0.3, and Llama 3.1 8B --- form a single high-correlation block. Across the ten pairwise comparisons among the five mature instruct models, the Spearman $\rho$ values range from 0.742 (Qwen 2.5 $\times$ Llama 3.1) to 0.919 (Llama 3.2 $\times$ Llama 3.1), with a mean of 0.834. The corresponding base-model block shows a comparable range and mean, and base-instruct pairs within each family are at the top of the distribution and all exceed $\rho$ = 0.91: Mistral-I $\times$ Mistral-B = 0.985, Qwen-I $\times$ Qwen-B = 0.975, Llama 3.1-I $\times$ Llama 3.1-B = 0.950, SmolLM2-I $\times$ SmolLM2-B = 0.922, Llama 3.2-I $\times$ Llama 3.2-B = 0.918.

Three things are worth noting about this matrix. First, every pair among the five mature architectures correlates above $\rho$ = 0.74, and this floor holds even across the largest architectural gap in the dataset (Qwen 2.5 1.5B's GQA-style decoder vs Llama 3.1 8B's grouped-attention 32-layer transformer) and the largest scale gap (1.5B vs 8B). The 21-emotion geometry is therefore not the property of any one architecture, training pipeline, or scale; it is shared across all five mature SLMs at correlations that would be unusually high even for repeated runs of the same model under different random seeds.

Second, the within-family pairs sit at the top of the distribution, both in instruct-instruct form (Llama 3.2 $\times$ Llama 3.1 = 0.919, the single highest cross-family $\rho$ in the matrix) and in base-instruct form within each family (mean $\rho$ = 0.927 across the five within-family base-instruct pairs). Family identity is therefore visible at the emotion-geometry level even when scale is held constant and even when RLHF is applied. We develop the within-family scale story further in Section 3.4.

Third, Gemma-3 1B is conspicuously outside this block. Its instruct variant correlates with the five mature models at $\rho$ between 0.527 and 0.625 --- substantially weaker than any mature-mature pair --- and its base variant correlates with everything in the matrix (including its own instruct version, $\rho$ = 0.190) at near-zero levels ($\rho$ between 0.181 and 0.252 across all eleven other models). We treat Gemma-3 not as a counter-example to the universality finding but as a separate maturity-regime case, and we develop it in Section 3.3.

The universality result in this section is the central empirical claim of the paper. The size of the correlations is what makes it surprising: across an order of magnitude in parameter count and across architectural choices that differ in attention pattern, normalization, tokenizer, and training data, the cosine geometry of 21 emotions is essentially the same geometry, up to noise that is comparable to within-family base-vs-instruct noise. Whatever process places these 21 concepts in residual stream space appears to converge to a similar arrangement under most training regimes that successfully produce a usable language model.

\begin{figure}[H]
\centering
\setcounter{figure}{0}
\includegraphics[width=\linewidth]{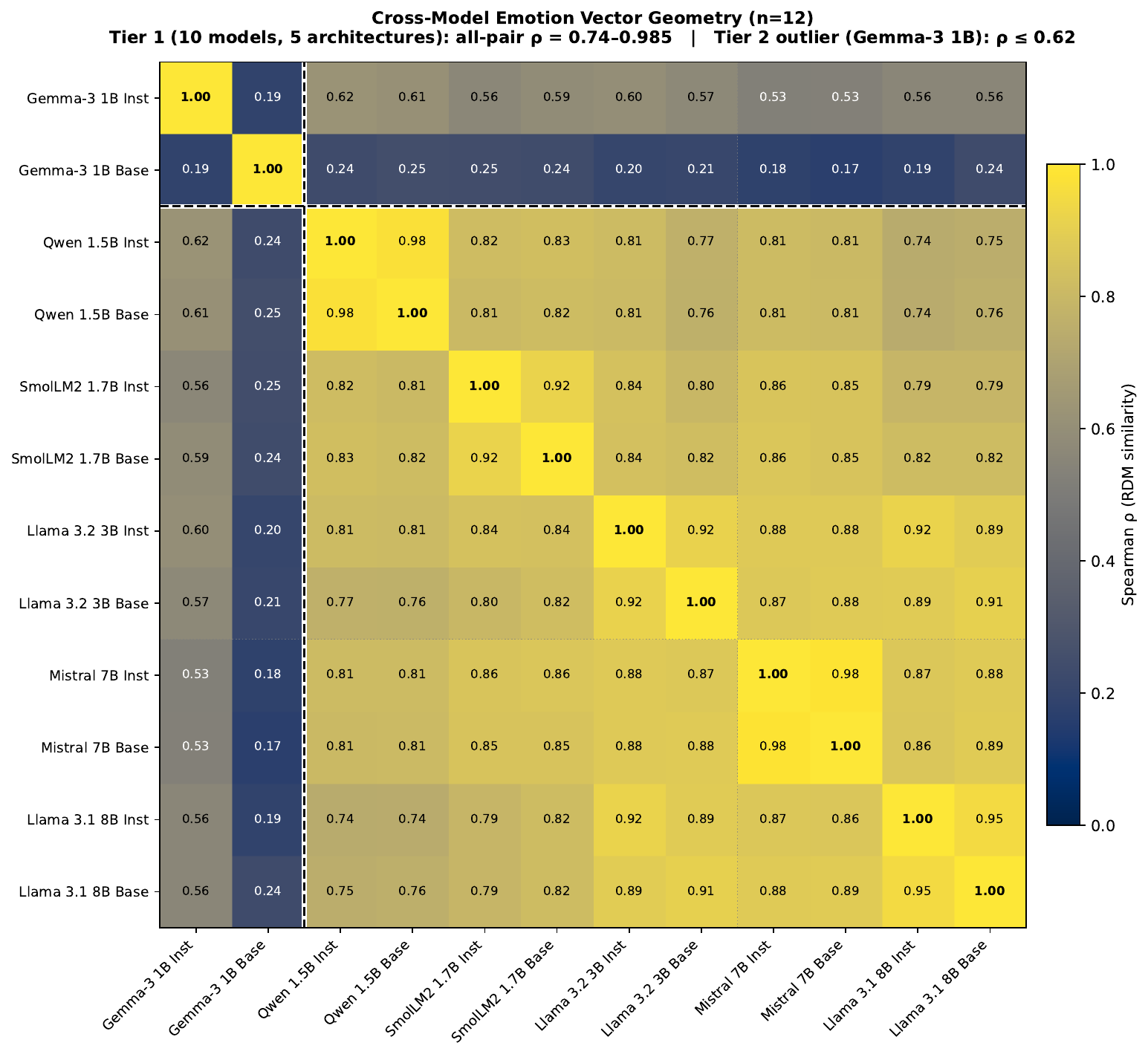}
\caption{\textit{Cross-model 21-emotion vector geometry across twelve small language models.} 12 $\times$ 12 RDM-of-RDMs matrix. Each cell shows the Spearman correlation between the off-diagonal entries of two models' 21 $\times$ 21 cosine RDMs at their respective best layers (Section 2.5). Models are grouped by family with base and instruct variants adjacent. The five mature architectures form a single high-correlation block ($\rho$ = 0.74--0.92 among instruct pairs), and base $\times$ instruct pairs within each family are the strongest alignments in the matrix: Mistral 7B v0.3 $\rho$ = 0.985, Qwen 2.5 $\rho$ = 0.975, Llama 3.1 $\rho$ = 0.950, SmolLM2 $\rho$ = 0.922, Llama 3.2 $\rho$ = 0.918. Gemma-3 1B base, with anisotropy 0.997, sits as an isolated noise-floor row ($\rho$ = 0.17--0.25 to all other models) regardless of architectural similarity; its instruct variant recovers to $\rho$ = 0.53--0.63 against the mature block but does not reach the mature-architecture range. See Section 3.1 for substantive interpretation and Section 4.1 for the anisotropy reliability discussion.}
\label{fig:1}
\end{figure}

\subsection{Behavioral--representational dissociation: Qwen vs Llama 3.2 on Compliance}

A natural follow-up question is whether models that \textit{behave} very differently also \textit{represent} emotions differently. If the universality result of Section 3.1 reflected a shallow, behavior-tracking property of the residual stream, we would expect models with sharply opposed behavioral profiles to show correspondingly weakened geometric alignment.

Before we present the test, a terminological note is in order. The MTI "Compliance" axis we use here is \textit{not} the same construct as the safety/refusal sense of "compliance" that has become standard in AI safety work, where it refers to whether a model accepts or refuses harmful prompts. The MTI Compliance axis measures, at a more general level, the model's responsiveness to social and procedural pressure on neutral content --- its tendency to flip its stated factual beliefs under repeated user pushback (facet B) and its tendency to actually carry out stylistic constraints when explicitly asked to (facet D), as detailed in Section 2.7. Neither facet involves harmfulness, jailbreak resistance, or refusal of disallowed requests at all. The two notions of compliance are related conceptually but measure different things, and the dissociation we report in this section is about MTI Compliance, not about safety compliance.

We test the question directly using the Qwen 2.5 1.5B Instruct vs Llama 3.2 3B Instruct pair, which the MTI Compliance facets identify as occupying opposite poles on both facets simultaneously (Section 2.7). Qwen scores B = 0.80 / D = 0.61 (epistemically yielding under user pushback, only moderately compliant with stylistic constraints among the format tasks it can perform at all), while Llama 3.2 scores B = 0.20 / D = 0.85 (epistemically firm under pushback, highly compliant with stylistic constraints). The two facets pull in opposite directions for these two models in the colloquial sense of "compliance": Qwen complies with social pressure to revise its claims but does not comply with format requests, while Llama 3.2 does the reverse. This is the cleanest within-dataset case of opposed behavior we have.

We computed the cosine RDM Spearman correlation between these two models at their respective best layers (Qwen layer 15 of 28, Llama 3.2 layer 11 of 28). The result is \textbf{$\rho$ = 0.813} (Pearson r = 0.840), well within the mature-architecture distribution of Section 3.1 (0.742--0.919) and in fact slightly above its mean. We do not report a parametric p-value here because the off-diagonal RDM entries are not independent --- each of the 21 emotion vectors participates in 20 pairwise comparisons --- and standard parametric significance tests on Spearman correlations between RDMs assume independence of the underlying samples. The substantive interpretation rests on the effect size and on the comparison to the within-dataset null distribution of Section 3.1, not on a p-value. The two models' 21-emotion geometries are statistically indistinguishable from any other mature-architecture pair in our dataset, even though their behavioral Compliance profiles are inverted on both facets that the MTI assesses.

The implication is that behavioral facet differences of the kind MTI is designed to detect are not written into the emotion representation itself. Whatever shapes the geometry of the 21 emotion vectors is upstream of, and largely independent from, whatever shapes a model's tendency to flip under social pressure or to follow format constraints. The latter must therefore arise at a downstream layer that uses this shared substrate --- for example, in the policy that decides whether to act on the contents of the representation, in the role-play and persona conditioning produced by RLHF, or in token-level generation dynamics that we are not measuring here. Mechanistically locating that downstream layer is outside the scope of this paper, but the dissociation itself is a strong constraint on any account of model behavior that would predict tight coupling between a model's "personality" (in the temperament sense) and its low-level affective representation.

We note that this is a single-pair test, and we are not claiming that all behavioral differences fail to appear in emotion geometry. We are claiming that the strongest within-dataset behavioral contrast we have access to does not appear, which is sufficient to motivate treating the universality result of Section 3.1 as something more than a surface artifact.

\begin{figure}[H]
\centering
\setcounter{figure}{1}
\includegraphics[width=\linewidth]{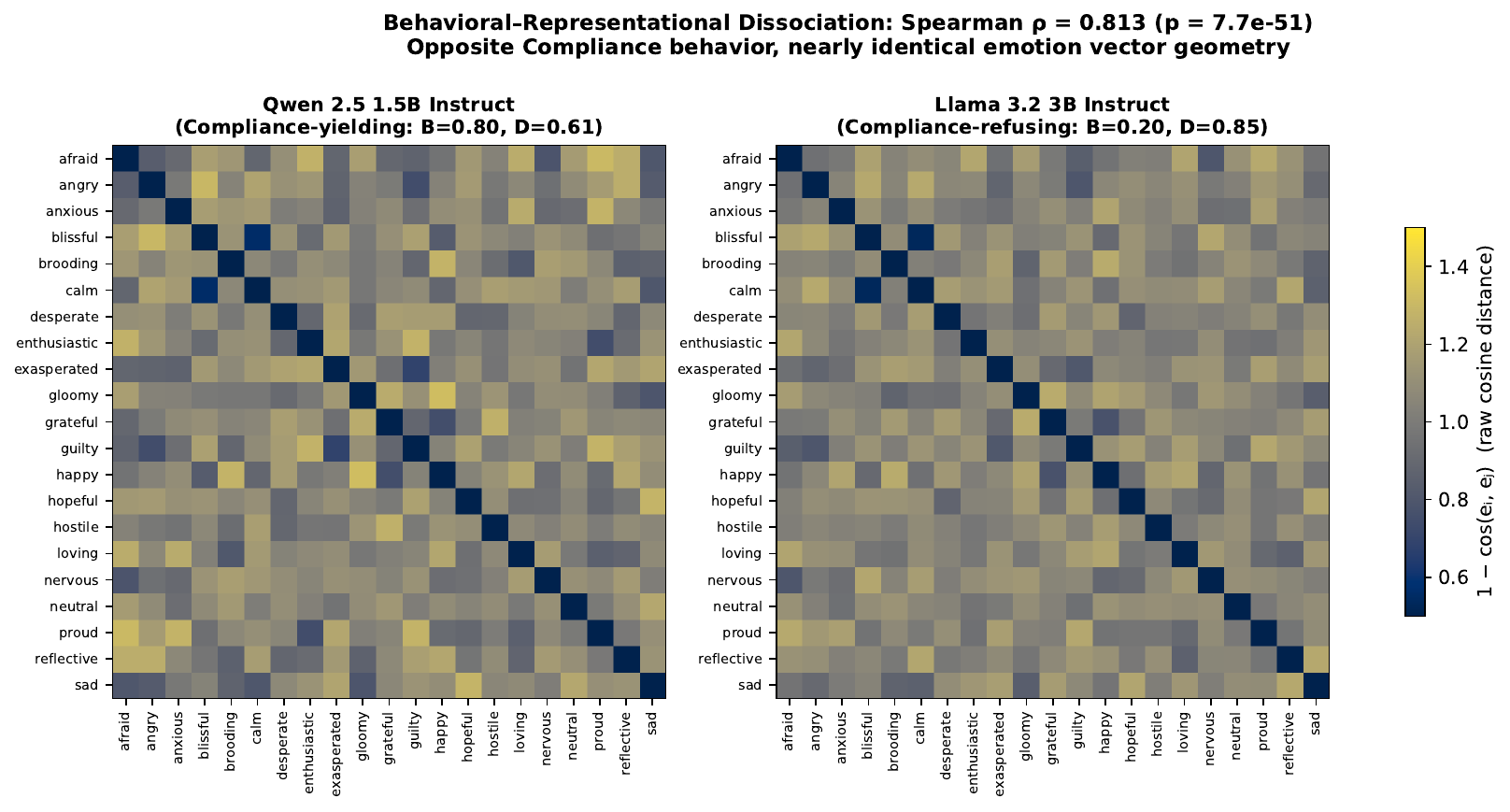}
\caption{\textit{Behavioral--representational dissociation: Qwen 2.5 $\times$ Llama 3.2 case study.} Side-by-side raw 21 $\times$ 21 cosine distance RDMs (rows and columns ordered alphabetically by emotion label) for Qwen 2.5 1.5B Instruct (left) and Llama 3.2 3B Instruct (right), each computed at the respective model's best layer (Qwen layer 15 of 28, anisotropy 0.848; Llama 3.2 layer 11 of 28, anisotropy 0.627). The two models occupy opposite poles of the MTI Compliance facets: Qwen scores 0.80 on the epistemic facet (B) and 0.61 on the procedural facet (D), an overall yielding profile; Llama 3.2 scores 0.20 on B and 0.85 on D, an inverted profile --- epistemically firm but procedurally compliant. Despite this behavioral inversion, the two RDMs are visually and statistically near-identical: the upper-triangle Spearman correlation between them is $\rho$ = 0.813 (Pearson r = 0.840), well within the mature-architecture distribution of Section 3.1 (0.742--0.919) and slightly above its mean of 0.834. The behavioral facet differences therefore arise above, not within, the shared emotion representation --- a negative result that constrains any account of compliance behavior that would tie it tightly to the low-level affective substrate. Raw cosine distance is used without anisotropy normalization (see Section 4.1 for the rationale). See Section 3.2 for the full analysis.}
\label{fig:2}
\end{figure}

\subsection{Representation maturity and the differential effect of RLHF}

Gemma-3 1B is the lowest-scale model in our dataset and also the most informative one for understanding how emotion geometry comes into existence. Its base variant (Gemma-3 1B PT) exhibits the largest geometric anomalies in the entire dataset: anisotropy at the best layer is 0.997 (the residual stream is essentially one-dimensional in cosine terms), the 21 $\times$ 21 RDM standard deviation is 0.333 (nearly three times the next-highest value in the dataset), and beyond approximately layer 12 the per-layer mean cosine becomes numerically NaN under fp16, indicating that the residual stream is not just degenerate but actively unstable in the model's deeper half. The best layer (layer 9 of 26, 34.6\% depth) sits in the still-stable lower half of the network. Steering at this layer produces an \textit{explosive} regime: small interventions yield empty completions and larger ones produce repetition collapse into single words ("contemplation contemplation contemplation\ldots{}", "assault assault assault\ldots{}").

We note one methodological caveat that will recur in the discussion (Section 5.3): under fp16, Gemma-3 1B base becomes numerically unstable beyond approximately layer 12, and we cannot fully separate how much of its extreme anisotropy reflects a genuine representational property from how much reflects precision artifact. The best layer (layer 9) sits within the stable region, so the descriptors we report are well-defined numerically; but a bf16 or fp32 re-extraction of Gemma-3 1B base would be a natural robustness check, and we flag this as an open question rather than treating the 0.997 / 0.333 numbers as definitive characterizations of the model's representational capacity.

The instruct variant (Gemma-3 1B IT) is qualitatively different. Anisotropy at best layer drops modestly from 0.997 to 0.982, but RDM standard deviation drops from 0.333 to 0.165 --- a 50\% reduction. The best layer shifts upward from layer 9 to layer 15 (34.6\% $\rightarrow$ 57.7\% of total depth), almost the full width of the network. The steering regime moves from \textit{explosive} to \textit{surgical}: interventions at low strengths produce coherent emotion-conditioned text ("It's a quiet feeling, a stillness that settles deep within me\ldots{}") and only collapse into repetition or off-language tokens at the highest strengths. All four geometric descriptors that we computed for this model change substantially between base and instruct.

For comparison, the same four descriptors barely move between base and instruct in any of the other five families. RDM standard deviation changes by less than 0.01 absolute in Qwen 2.5, Llama 3.2, Mistral 7B, and Llama 3.1 8B; by 0.005 in SmolLM2. Best-layer position is identical in Qwen, Llama 3.2, and Mistral 7B (v0.3 base and v0.3 instruct both select layer 13 of 32 = 40.6\% depth), and shifts by less than 5 percentage points in Llama 3.1. The largest non-Gemma RLHF-induced shift is in SmolLM2 (best layer 54.2\% $\rightarrow$ 45.8\%), and even there the underlying RDM std and anisotropy are essentially unchanged. The five mature within-family base $\times$ instruct RDM correlations reflect this uniformly: Mistral 7B $\rho$ = 0.985, Qwen 2.5 $\rho$ = 0.975, Llama 3.1 8B $\rho$ = 0.950, SmolLM2 $\rho$ = 0.922, Llama 3.2 $\rho$ = 0.918 --- all above 0.91, and the two 7B+ models (Mistral 7B and Llama 3.1 8B) showing the two highest within-family alignments in the mature set apart from Qwen. None of the five mature families show anything resembling Gemma-3's RDM std reduction or its 23-percentage-point best-layer shift.

The compact statement of this finding is best framed as a strong hypothesis rather than as an established result, since it rests on a single immature-regime case in our dataset: \textbf{RLHF appears to have substantial geometric effects primarily on representations that have not yet matured}. For the five mature families whose base models already exhibit anisotropy below 0.86 and RDM std below 0.13 (Qwen 2.5, SmolLM2, Llama 3.2, Mistral 7B, Llama 3.1), RLHF leaves emotion geometry essentially unchanged. For Gemma-3 1B, whose base model has not yet differentiated emotions in any meaningful sense (RDM std 0.333, anisotropy 0.997, post-best-layer numerical instability), RLHF acts on every descriptor simultaneously and shifts the model from an explosive to a surgical steering regime. Within this single contrast, mature models appear not to need RLHF to have an emotion geometry, and the one immature model we have appears to acquire most of its geometry from post-training.

This pattern is consistent with --- and would add geometric evidence to --- accounts in the literature that frame instruction tuning and RLHF as primarily structuring or exposing capabilities already present in the pretrained base, rather than installing new representations from scratch (Zhou et al. 2023; Lin et al. 2024), if it generalizes. It also yields a concrete prediction: smaller models below some maturity threshold should show large RLHF effects on emotion geometry, while larger or otherwise more mature models should show small effects regardless of how much post-training they receive. We do not have enough models in the immature regime (Gemma-3 1B is the only one) to test this prediction across families, and we are explicit throughout the rest of the paper that this is a hypothesis the present work raises rather than settles. The Section 5.3 discussion returns to the n = 1 caveat in more detail.

\subsection{Family identity preservation across scale}

The single highest cross-family RDM correlation in the entire dataset is between Llama 3.2 3B Instruct and Llama 3.1 8B Instruct, at $\rho$ = 0.919 --- higher than any other instruct-instruct pair, including pairs from architecturally closer families. The corresponding base pair (Llama 3.2 3B Base $\times$ Llama 3.1 8B Base) is $\rho$ = 0.910. Both are at the top end of the mature-architecture distribution.

What makes this notable is the scale gap. Llama 3.1 8B has 2.7$\times$ the parameters of Llama 3.2 3B, with the same number of layers (32 vs 28) but a 33\% larger residual stream (4096 vs 3072). The two models also share architectural lineage but were trained at different times on different data mixtures. Despite these differences, their emotion vector geometries are more closely aligned with each other than either is with any non-Llama mature model in the dataset.

The geometric descriptors tell a related story. Best-layer position is 39.3\% for Llama 3.2 3B and 37.5\% for Llama 3.1 8B --- within two percentage points despite the model size difference and despite Llama 3.1 having four more layers in absolute terms. Best layer percentage is therefore stable along the Llama line in a way that it is not stable across families more generally (the full dataset spans 25\% to 58\%, see Section 3.5). RDM standard deviation is 0.103 vs 0.106; anisotropy is 0.627 vs 0.680. Within the noise of these measurements, the two Llama models look like the same model at two scales rather than like two members of a heterogeneous family.

We read this as evidence that family identity --- the choices of architectural family, training data lineage, and training pipeline --- leaves a measurable signature on emotion geometry that is preserved across scale within the family. The signature is strong enough to dominate scale effects within the within-family contrast we have. Whether this generalizes to other families with multiple available scales (e.g. Qwen 2.5 has multiple sizes we did not include) is a question for follow-up work; for the present paper, the Llama 3.2 $\times$ Llama 3.1 pair is one informative data point and is the strongest evidence in the matrix that emotion geometry tracks model lineage and not just emergent properties of "language model trained at scale X".

\begin{figure}[H]
\centering
\setcounter{figure}{3}
\includegraphics[width=\linewidth]{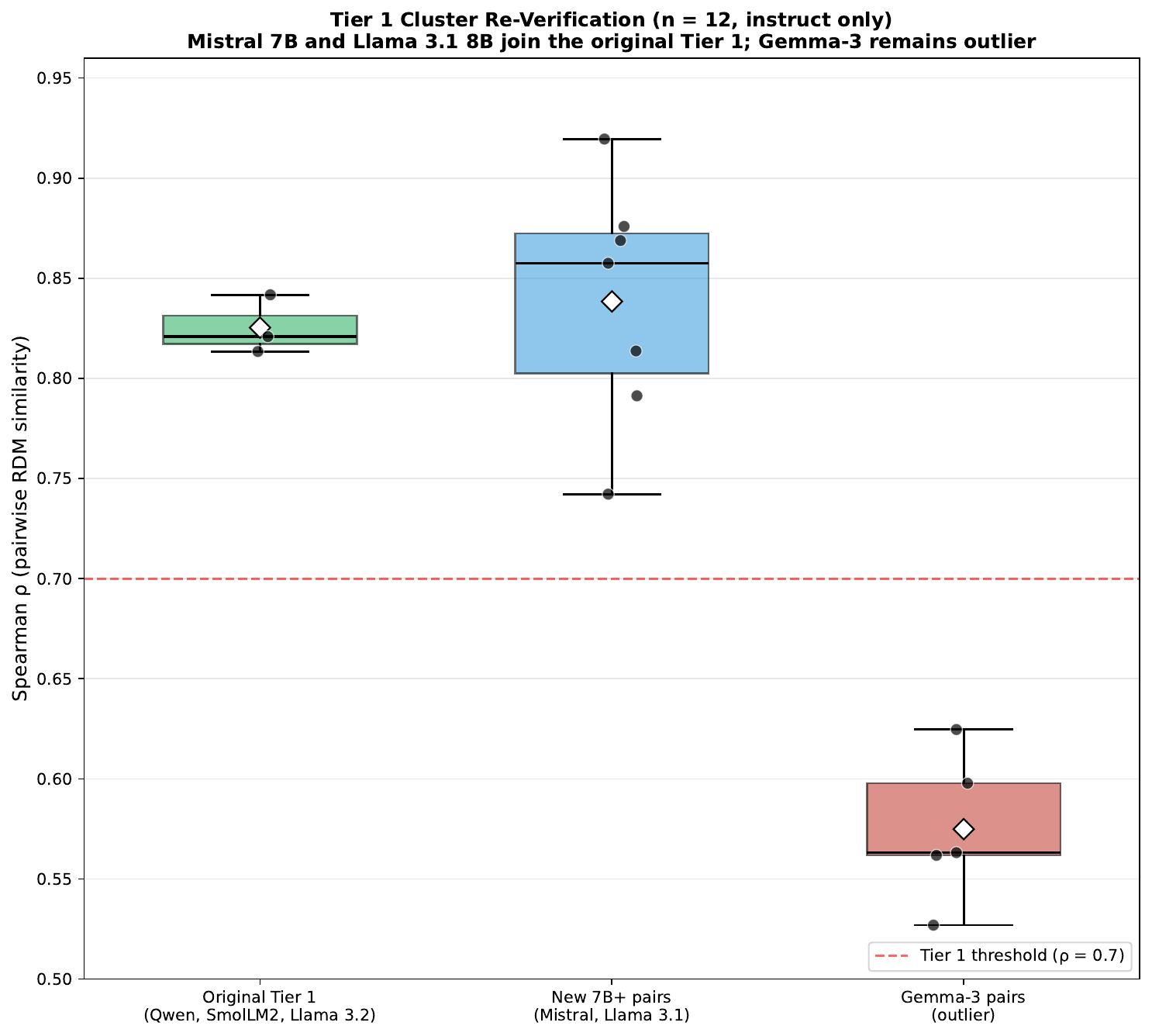}
\caption{\textit{Tier 1 cross-model RDM alignment by pair group (instruct-only).} Boxplot of pairwise RDM Spearman correlations across the C(6,2) = 15 instruct model pairs in our dataset, grouped by the size class and family composition of each pair. \textbf{Original Tier 1} (n = 3) comprises the three sub-4B-parameter mature pairs --- Qwen 2.5 $\times$ SmolLM2, Qwen 2.5 $\times$ Llama 3.2, SmolLM2 $\times$ Llama 3.2 --- with mean $\rho$ = 0.825 (range 0.813--0.842). \textbf{New 7B+ pairs} (n = 7) comprises all instruct pairs involving either Mistral 7B v0.3 or Llama 3.1 8B, contributed by this paper's addition of 7B+ models to the mature set, with mean $\rho$ = 0.838 (range 0.742--0.919). The maximum of this group ($\rho$ = 0.919) is the Llama 3.1 8B $\times$ Llama 3.2 3B cross-generation within-family pair, consistent with the family-identity result of Section 3.4. \textbf{Gemma pairs} (n = 5) comprises the pairs between Gemma-3 1B instruct and each of the five mature instruct models, with mean $\rho$ = 0.575 (range 0.527--0.625) --- substantially below both mature groups, consistent with Gemma-3's still-partial recovery from its immature base form (Section 3.3). The overlap between the Original Tier 1 distribution and the New 7B+ distribution shows that adding 7B+ models to the mature set does not pull the distribution upward or downward, and establishes that the universality result of Section 3.1 is not specific to one size range within the mature regime. The horizontal reference line at $\rho$ = 0.7 marks the Tier 1 inclusion threshold; all 10 mature-mature pairs fall above it and all 5 Gemma pairs fall below. Box elements: IQR with median; diamond: group mean; individual points: single pair values.}
\label{fig:4}
\end{figure}

\subsection{Size--maturity correlations across the twelve models}

Sections 3.3 and 3.4 raise the question of whether the geometric descriptors of Section 2.5 show systematic dependence on model size when the dataset is treated as a continuous distribution rather than as a discrete set of cases. We computed Spearman correlations across all twelve models between size (in billions of parameters) or residual-stream width (\texttt{d\_\allowbreak{}model}) on one axis and each of the geometric descriptors on the other. Figure 3 visualizes the scatter plots underlying the correlations reported in this section.

The headline numbers are:

\begin{table}[H]
\centering
\footnotesize
\begin{tabular}{llll}
\toprule
Predictor & Outcome & Spearman $\rho$ & p (uncorrected) \\
\midrule
\texttt{d\_\allowbreak{}model} & anisotropy at best layer & \textbf{$-$0.882} & 0.0001 \\
size (B) & anisotropy at best layer & $-$0.820 & 0.0011 \\
size (B) & RDM standard deviation & \textbf{$-$0.891} & 0.0001 \\
size (B) & best layer \% & $-$0.484 & 0.11 \\
\bottomrule
\end{tabular}
\end{table}

Anisotropy and RDM standard deviation both show strong monotonic relationships with model size. At n = 12 the effect sizes ($\rho$ $\approx$ $-$0.88 to $-$0.89 for both) are the primary evidence; the associated parametric p-values are reported in the table for completeness but are not corrected for the multiple tests shown and should not be read as the primary support for the claim. Under a Bonferroni correction ($\alpha$ = 0.05/4 = 0.0125) the two top rows remain significant and the best-layer-\% row does not, which is consistent with the effect-size reading. We also note that \texttt{d\_\allowbreak{}model} and parameter count are themselves nearly collinear across our dataset, so the two anisotropy rows in the table are not independent tests; the choice between them as predictor is essentially cosmetic, and results are essentially identical when \texttt{d\_\allowbreak{}model} is used as the size predictor for RDM std and best-layer \% as well (the largest difference across the four tests is 0.06 in $\rho$).

The relationship is not subtle: Mistral 7B base sits at anisotropy 0.532 and RDM std 0.097, while Gemma-3 1B base sits at anisotropy 0.997 and RDM std 0.333. The three mature small-to-mid families (Qwen 1.5B, SmolLM2 1.7B, Llama 3.2 3B) sit between, monotonically closer to the Mistral/Llama 3.1 end than to the Gemma end. The "maturity gradient" of Section 3.3 is not a Gemma-vs-everyone-else dichotomy: it is a continuous gradient on which Gemma-3 1B happens to be the lowest point and 7B--8B models the highest.

Best-layer percentage tells a different story. Its correlation with size is much weaker ($\rho$ = $-$0.484) and does not reach conventional significance at n = 12 (p = 0.11). Inspecting the per-model values reveals why: best-layer percentage clusters within families (Qwen \textasciitilde{}54\%, both Llama generations \textasciitilde{}38\%, Mistral 7B \textasciitilde{}41\%, Gemma instruct \textasciitilde{}58\%, SmolLM2 \textasciitilde{}50\%) and the two 7B+ models converge near 38--41\% of total depth rather than being pulled further toward the output end by scale. Best-layer percentage is therefore better understood as a family signature (Section 3.4) than as a function of scale per se. Where in the network the optimal extraction depth sits is a property the family carries with it, while how degenerate or differentiated the residual stream is at that depth is a property that scales with size.

\begin{figure}[H]
\centering
\setcounter{figure}{2}
\includegraphics[width=\linewidth]{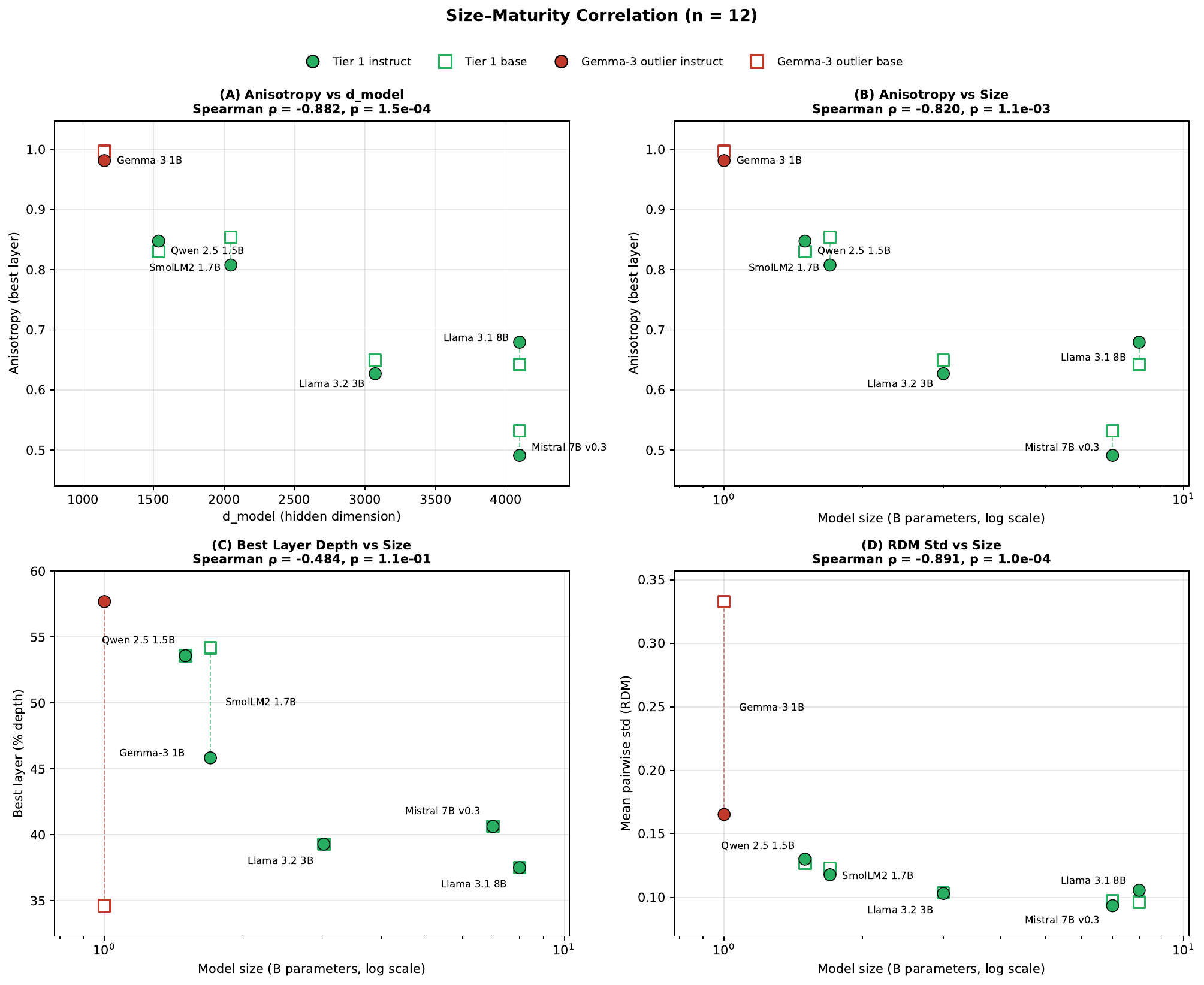}
\caption{\textit{Representation maturity gradient scales with model size (n = 12).} Scatter plots showing three of the four size--geometry correlations of Section 3.5, with each of the twelve models plotted as a single point. (a) Anisotropy at best layer vs hidden dimension \texttt{d\_\allowbreak{}model}: Spearman $\rho$ = $-$0.882, p = 0.0001. (b) RDM standard deviation at best layer vs parameter count (B): $\rho$ = $-$0.891, p = 0.0001. (c) Best-layer percentage (index as a fraction of total depth) vs parameter count: $\rho$ = $-$0.484, p = 0.11 (not significant at n = 12). Anisotropy and RDM standard deviation --- the "sharpness" dimensions of the representation --- scale monotonically with size: larger models have less degenerate residual streams and more differentiated emotion structure, with Mistral 7B v0.3 and Llama 3.1 8B at the mature end and Gemma-3 1B base as the clear outlier at anisotropy 0.997. Best-layer percentage, by contrast, clusters within families (Qwen \textasciitilde{}54\%, Llama 3.2 and 3.1 \textasciitilde{}37--39\%, Mistral 7B \textasciitilde{}41\%, SmolLM2 \textasciitilde{}45--54\%, Gemma instruct \textasciitilde{}58\%) rather than shifting systematically with scale, indicating that location is a family signature while sharpness scales with size. See Section 3.5 for the size-correlation analysis and Section 3.4 for the family-signature interpretation of the best-layer plot.}
\label{fig:3}
\end{figure}

The differentiated picture that emerges from the four correlations is informative. Scale matures emotion geometry along the \textit{sharpness} dimension (anisotropy down, RDM std up in absolute terms --- i.e. the residual stream becomes less degenerate and the 21 emotions become more differentiated from one another) but not along the \textit{location} dimension (where in the network the best emotion direction sits). Two different things are happening as models grow: their representations spread out and differentiate, and their families preserve where in the stack they place the emotion structure. The maturity gradient of Section 3.3 and the family identity signature of Section 3.4 are two consequences of this same picture, and the size correlations of this section make both of them quantitative.

These five findings together give the substantive result of the paper: a shared 21-emotion geometry across five mature SLM architectures, robust to the strongest behavioral contrast we can construct, sitting on top of a continuous maturity gradient that scales with model size in the sharpness dimension but is family-determined in the location dimension, with RLHF having a substantial geometric effect only at the immature end of this gradient. Section 4 turns to the methodological findings that constrain how this picture can be safely interpreted across studies.

\section{Results: Methodological Findings}

The previous section established the substantive picture: a shared 21-emotion geometry across five mature SLM architectures, behavioral--representational dissociation, a maturity gradient, family identity preservation, and size correlations. All of those results were computed on raw cosine RDMs from a single unified comprehension extraction pipeline. This section presents the three methodological findings that justify those choices and that constrain how cross-model SLM emotion comparisons can be done safely going forward. Section 4.1 addresses the anisotropy problem and explains why the natural attempt to "normalize it away" does not work mathematically. Section 4.2 documents the cross-backend equivalence test that licenses our mixed-backend dataset. Section 4.3 develops the four-layer decomposition of method $\times$ precision conflation, which is the most consequential of the three for reading the prior literature.

\subsection{The anisotropy reliability problem and the invariance of linear normalization}

Across our twelve models, anisotropy at the best layer ranges from 0.491 (Mistral 7B instruct) to 0.997 (Gemma-3 1B base) --- a span that covers nearly the entire theoretically possible range. The largest cluster sits between roughly 0.49 and 0.85, with Mistral, both Llamas, and Qwen / SmolLM2 distributed within it; Gemma-3 1B sits as an isolated outlier far above the cluster, with both base (0.997) and instruct (0.982) above the 0.95 mark. This spread raises an obvious comparability question: when one model in a pair has anisotropy 0.85 and the other has 0.997, can their cosine RDMs be meaningfully compared at all, and if so, how?

\subsubsection{Empirical observation: high-anisotropy outliers cluster at low $\rho$}

The empirical pattern in our 12 $\times$ 12 RDM-of-RDMs matrix is unambiguous. Gemma-3 1B base correlates with the eleven other models in our dataset at Spearman $\rho$ between 0.171 and 0.252, with no clear dependence on family, scale, or architectural similarity (Llama 3.2 base $\rho$ = 0.212; Llama 3.1 base $\rho$ = 0.236; Qwen base $\rho$ = 0.252; SmolLM2 base $\rho$ = 0.241; Mistral base $\rho$ = 0.171; Gemma-3 instruct $\rho$ = 0.190). The eleven correlations are tightly clustered around 0.21 with a spread of only $\pm$0.04 --- the kind of pattern one expects when the cosine geometry of one of the two models is dominated by floating-point and tokenizer noise rather than by signal. By contrast, Gemma-3 1B instruct, whose anisotropy has dropped from 0.997 to 0.982, correlates with the same eleven models in the range 0.527--0.625 --- still well below the mature-architecture distribution of 0.74--0.92, but far above its own base form, and showing the within-family signature one would expect from a real emotion geometry (the strongest of the eleven correlations is with its own base variant only at $\rho$ = 0.190, which is part of the same noise-floor pattern).

The contrast between Gemma-3 base and Gemma-3 instruct is the cleanest illustration in the dataset that \textbf{anisotropy controls the reliability of the cosine RDM, not just its scale}. Two RDMs computed from very high-anisotropy residual streams will show low correlation with everything, including with each other and with their own family members, because the RDM entries themselves are noise-dominated. Treating Gemma-3 base's $\rho$ $\approx$ 0.20 with everyone as a substantive finding about Gemma-3's emotion representation would therefore be a measurement error: the model has too little geometric signal at the residual stream to support a cosine-based comparison at all.

\subsubsection{The mathematical point: per-RDM linear normalization is invariant under rank correlation}

The natural first response to this problem is to apply a per-RDM normalization that adjusts for the model's anisotropy. A simple form is

RDM' = (RDM $-$ (1 $-$ anisotropy)) / std(RDM),

which centers each RDM at its own anisotropy-derived baseline and rescales by its own spread. This is the operation we initially implemented in our pilot analysis script. It is also, we now observe, mathematically a no-op for the cross-model similarity numbers we care about.

The reason is that Spearman $\rho$ depends only on the ranks of the RDM entries, and any monotonically increasing transformation preserves ranks. The transformation \texttt{x -> (x - a) /\allowbreak{} b} for any constants \texttt{a} and any \texttt{b > 0} is monotonic, so the ranks of the entries of \texttt{RDM'} are identical to the ranks of the entries of \texttt{RDM}. Therefore, for any two RDMs \texttt{R1} and \texttt{R2} with any per-RDM choices of \texttt{(a1, b1)} and \texttt{(a2, b2)},

Spearman(R$_{1}$', R$_{2}$') = Spearman(R$_{1}$, R$_{2}$).

Pearson correlation gives the same conclusion by a slightly different argument: under independent linear transformations of two variables, both the covariance and the product of the standard deviations rescale by the same factor \texttt{b1 * b2}, which cancels in the Pearson formula. Linear normalization therefore cannot change either Spearman $\rho$ or Pearson r between two RDMs, no matter how different the per-model anisotropies are. The cross-model correlation reported under "anisotropy normalization" with this kind of scheme is the same number as the raw cosine correlation, to within numerical error.

This is not a property of our specific normalization formula. Any normalization that operates linearly on each RDM independently will have the same invariance, including subtraction of any per-RDM baseline, division by any per-RDM scale, z-scoring, and any composition of these. To actually change the cross-model correlation, one would need a transformation that is either (a) non-linear in the entries of a single RDM, (b) defined over the underlying vectors before the cosine RDM is computed (for example, by projecting out the dominant residual-stream directions before computing cosines, in the spirit of the "All-but-the-Top" isotropization method of Mu et al. 2018), or (c) defined jointly over both RDMs in a way that changes ranks. None of these are routinely applied in current SLM emotion-vector work, and none of them is implemented in our analysis pipeline.

\subsubsection{Practical recommendation}

A reader might object at this point that the invariance result implies the entire enterprise of RDM-based comparison should be abandoned in favor of an absolute distance metric --- for example, Frobenius distance between the two RDMs treated as flat matrices, or a Wasserstein distance defined over the underlying vector distributions. Such metrics are not invariant under per-RDM rescaling, and in principle could distinguish "Gemma-3 base has a structurally different emotion geometry" from "Gemma-3 base has the same geometry but with smaller absolute spread." We do not pursue this route, for two reasons. First, our object of interest in Section 3 is the \textit{relational} structure of the 21 emotions --- which pairs are close to one another, which are far apart, and how the rank order of all 210 pairwise relationships compares between two models --- not the absolute geometric distance between two vector configurations in their respective residual stream geometries. Spearman correlation on the upper triangle of the cosine RDM is the natural metric for that question precisely because it asks about rank order rather than about absolute magnitude, and is therefore invariant to the kinds of per-model rescalings that are not part of the question we want to answer. Second, an absolute-distance metric would answer a different and orthogonal question --- how far apart, in some chosen geometry, are two configurations? --- for which we have no comparable cross-model baseline and which would itself require new methodological infrastructure (an alignment scheme between residual streams of different dimensionality, a principled choice of geometry, a way to interpret what "distance N" means at a substantive level). Both routes are defensible; they answer different questions, and the present paper is about the relational-structure question.

Given that the most natural form of anisotropy correction within the relational-structure framing does nothing, and that more aggressive forms have not been validated in this literature, the correct response in our view is not to normalize but to flag. Specifically, we recommend that any cross-model SLM emotion-vector study report raw cosine RDM similarity \textit{together with the per-model anisotropy at the extraction layer}, so that readers can identify pairs in which one or both models have anisotropy above \textasciitilde{}0.95 and treat the corresponding similarity values as unreliable rather than as substantive. In the present paper, we follow this recommendation throughout: every $\rho$ value reported in Section 3 is raw cosine, and Gemma-3 1B base --- the only model in our dataset with anisotropy above the 0.95 reliability threshold at every reference layer (Section 2.5) --- is excluded from the universality claim of Section 3.1 and treated separately as part of the maturity-gradient story of Section 3.3. Gemma-3 1B instruct, whose anisotropy has dropped to 0.982, is borderline by this criterion; we report its cross-model correlations alongside its anisotropy and note that they fall between Gemma base and the mature cluster, consistent with the maturity-gradient interpretation.

A genuine isotropization analysis --- one that operates on the underlying vectors rather than on the RDM entries --- would be a valuable extension of this work and would let us re-examine whether Gemma-3 1B base has \textit{any} recoverable emotion structure once its degenerate residual stream is corrected for. We do not pursue it here, because the substantive findings of Section 3 do not require it: they are stable under raw cosine, and the role of Gemma-3 base in those findings is exactly the role its anisotropy says it should have, namely the lower endpoint of a maturity gradient.

\subsection{Cross-backend equivalence: TransformerLens and raw HuggingFace hooks extract the same vectors}

Eight of the twelve models in our dataset (Qwen 2.5 base/instruct, Llama 3.2 base/instruct, Gemma-3 base/instruct, Llama 3.1 base/instruct) were extracted via TransformerLens, which exposes the residual stream through named hook points. The remaining four (SmolLM2 base/instruct and Mistral 7B v0.3 base/instruct) were extracted via raw HuggingFace transformers hooks, because TransformerLens does not currently support either of those two architectures. Section 3 includes both backend groups in cross-model RDM comparisons against each other, so the substantive findings of the paper would be confounded if the two backends extracted materially different representations from nominally the same residual stream position.

We tested this directly. Llama 3.2 3B Instruct loads cleanly under both backends, so we ran the full comprehension-mode extraction pipeline on it twice --- once via TransformerLens and once via raw HuggingFace hooks --- and compared the resulting 21-emotion vector sets at the model's best layer (layer 11 of 28). The matched position is \texttt{blocks.11.\allowbreak{}hook\_\allowbreak{}resid\_\allowbreak{}post} in TransformerLens and \texttt{hidden\_\allowbreak{}states[12]} in HuggingFace (the +1 accounting for the input embedding occupying index 0 of the \texttt{hidden\_\allowbreak{}states} tuple).

Across all 21 emotions, the per-vector cosine similarity between the TL and HF extractions had mean 0.999998, minimum 0.999997, and maximum 0.999999. The 21 $\times$ 21 cosine RDMs computed from the two extractions had Spearman correlation 0.999990 and a relative Frobenius distance of 1.1 $\times$ 10$^{-4}$. These numbers are well within floating-point round-off range --- they are the kind of values one expects when two different computation paths arrive at the same answer modulo the order of operations in the underlying linear algebra kernels. We treat them as numerical equivalence.

The practical implication is that the cross-backend mix in our dataset is not a confound. Wherever Section 3 compares a TL-extracted model to an HF-extracted model --- for example, any of the pairs between \{SmolLM2 base/instruct, Mistral base/instruct\} and the eight TL models --- the comparison can be read as a direct cross-model comparison rather than as a confounded cross-backend comparison. We also note that this equivalence test is itself a useful result for any future SLM emotion-vector work that needs to mix the two backends --- for example, work that wants to use TransformerLens on architectures it supports while falling back to raw HuggingFace hooks on architectures it does not, which is the same situation we faced when SmolLM2 and Mistral 7B forced the four HF extractions in our dataset.

We caution that this test was run on a single model (Llama 3.2 3B Instruct) and so does not strictly establish that TL and HF agree for \textit{all} models on \textit{all} layers. What it establishes is that, at a matched position on a model that loads cleanly under both backends, the extracted vectors agree to within floating-point noise. We have no theoretical reason to expect this to fail on other models or other layers --- the residual stream at a given block boundary is the same tensor regardless of which library reads it --- and the test stands as a sanity check rather than as a comprehensive verification.

\subsection{Method $\times$ precision conflation: a four-layer decomposition}

The most consequential methodological finding of this paper concerns the apparent disagreement between comprehension- and generation-based emotion vector extraction, originally reported by Jeong (2026a). That paper extracted its two 7B+ models (Mistral 7B and Llama 3.1 8B) at INT8 precision under a VRAM constraint and its smaller models at fp16; the dissociation it highlighted was most prominent in the 7B+ subset, and the natural reading at the time was that the disagreement was a "method effect" --- comprehension and generation simply pick up different aspects of the model's affective representation. We now show that this reading is partially correct but conflates \textbf{three} distinct sources of variance --- method, sub-parameter choices within generation, and precision --- only one of which is the method per se, and that the naive cross-experiment statistic which pools these three sources distorts in opposite directions for different models. Below we develop these as three clean isolating contrasts (layers 1--3) plus a fourth contrast that reproduces the conflated cross-experiment statistic for comparison (layer 4). The decomposition was constructed by running the protocol-matched fp16 generation pipeline (\texttt{paper5\_\allowbreak{}extract\_\allowbreak{}protocol\_\allowbreak{}matched.\allowbreak{}py}, Section 2.3) on Mistral 7B Instruct and Llama 3.1 8B Instruct, the two models for which both our fp16 extractions and Jeong (2026a)'s INT8 generation extractions exist. Because these are also the two 7B+ models in Jeong (2026a)'s dataset, the precision contrast we develop below is specifically a contrast within the INT8-quantized subset of that earlier work; Jeong (2026a)'s smaller-model results, which were never quantized in the first place, are not affected by the precision confound we identify here.

We label four extraction conditions for the analysis below:

\begin{itemize}
\item \textbf{A}: comprehension-mode extraction at fp16 (this paper's main pipeline, applied to all twelve models in Section 3).
\item \textbf{B}: Jeong (2026a)'s generation pipeline at INT8, with its specific sub-parameter choices and quantized precision.
\item \textbf{C}: generation-mode extraction at fp16, with the eight protocol parameters held identical to Jeong (2026a)'s pipeline (the protocol-matched fork of Section 2.3) --- same generation method, same prompts, same sample count, same chat templating, same centering and normalization choices, but precision changed from INT8 to fp16.
\item \textbf{D}: generation-mode extraction at INT8 with protocol-matched parameters; this is operationally identical to \textbf{B} in our framework, and we use \textbf{D} and \textbf{B} interchangeably below to make the four-layer pairing visible.
\end{itemize}

The four contrasts that decompose the total method $\times$ precision effect --- three isolating contrasts plus one conflated cross-experiment statistic for comparison --- are:

\begin{table}[H]
\centering
\footnotesize
\begin{tabularx}{\linewidth}{llX}
\toprule
Layer & Contrast & What it isolates \\
\midrule
1 & A vs C & Method (comprehension vs protocol-matched generation), precision held at fp16 \\
2 & B vs C & Sub-parameter sensitivity within generation, method and precision held constant \\
3 & C vs D & Precision (fp16 vs INT8), generation method and sub-parameters held identical \\
4 & A vs D & The conflated cross-experiment statistic that mixes all three of the above (not an independent source of variance) \\
\bottomrule
\end{tabularx}
\end{table}

The four contrasts give very different pictures, and Figure 6 visualizes the same numbers as a grouped bar chart that makes the conflation in layer 4 visible at a glance:

\begin{table}[H]
\centering
\footnotesize
\begin{tabular}{llll}
\toprule
Layer & Contrast & Mistral 7B Instruct ($\rho$) & Llama 3.1 8B Instruct ($\rho$) \\
\midrule
1 & A vs C (clean method) & 0.091 & 0.355 \\
2 & B vs C (sub-parameter) & 0.019 & 0.033 \\
3 & C vs D (clean precision) & 0.215 & 0.527 \\
4 & A vs D (conflated) & 0.409 & 0.325 \\
\bottomrule
\end{tabular}
\end{table}

Each row tells a different story, and reading them together is the point of the analysis.

\begin{figure}[H]
\centering
\setcounter{figure}{4}
\includegraphics[width=\linewidth]{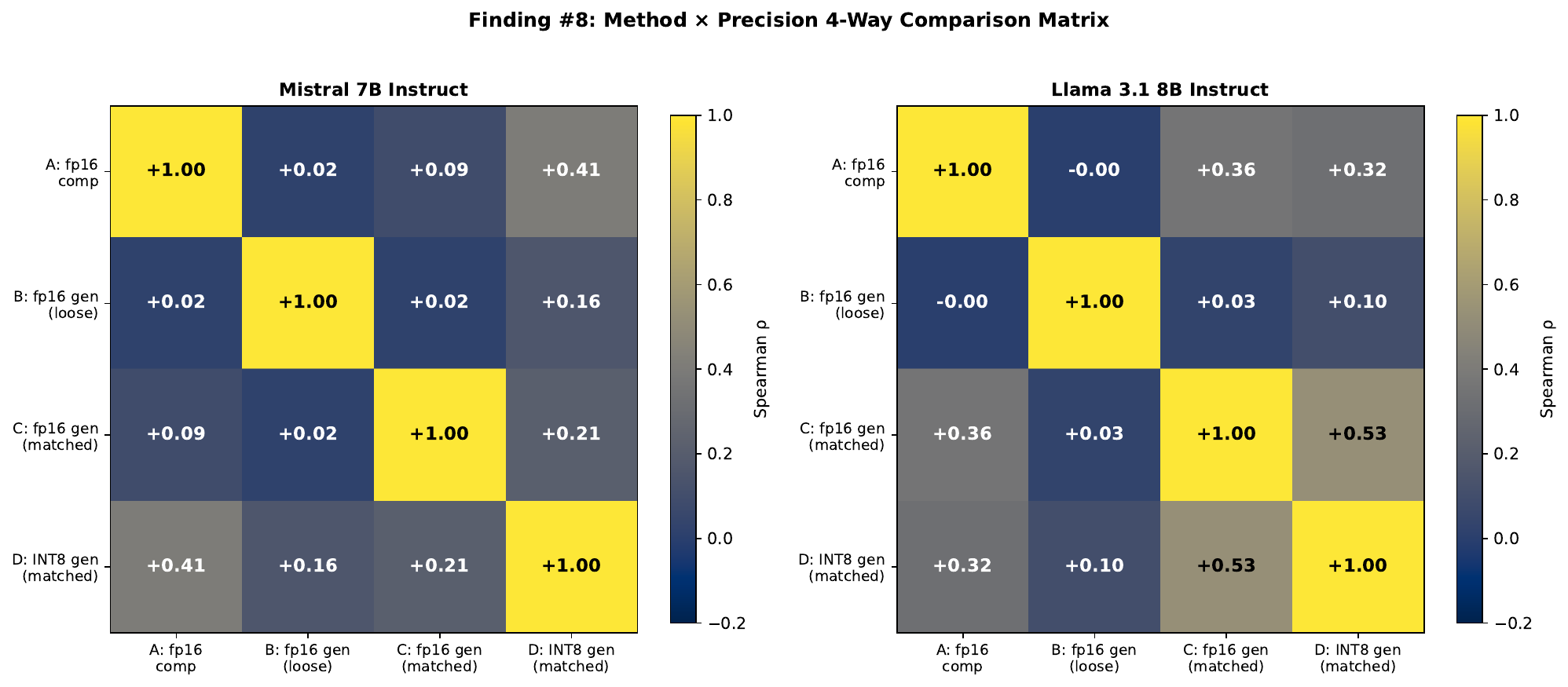}
\caption{\textit{Four-layer decomposition of the apparent comprehension-vs-generation method effect: 4-way pipeline comparison matrix.} For each of Mistral 7B Instruct (top row) and Llama 3.1 8B Instruct (bottom row), the 4 $\times$ 4 cell grid shows the Spearman correlation between the 21 $\times$ 21 cosine RDMs produced by four pipeline configurations: \textbf{A} = fp16 comprehension, \textbf{B} = fp16 generation with Jeong (2026a)'s original sub-parameter configuration, \textbf{C} = fp16 generation with the protocol-matched sub-parameter configuration (\texttt{paper5\_\allowbreak{}extract\_\allowbreak{}protocol\_\allowbreak{}matched.\allowbreak{}py}, Section 2.3), and \textbf{D} = INT8 generation as originally reported in Jeong (2026a). The four isolating contrasts are read off the off-diagonal cells: \textbf{A vs C} = method effect at fp16 with sub-parameters matched (Mistral $\rho$ = 0.091, Llama $\rho$ = 0.355); \textbf{B vs C} = sub-parameter sensitivity within fp16 generation (Mistral $\rho$ = 0.019, Llama $\rho$ = 0.033); \textbf{C vs D} = precision effect at protocol-matched generation (Mistral $\rho$ = 0.215, Llama $\rho$ = 0.527); \textbf{A vs D} = the conflated cross-experiment statistic originally reported for these models (Mistral $\rho$ = 0.409, Llama $\rho$ = 0.325). The conflated A--D cell is a mixture of the three preceding effects, not an estimator of any one of them, and in Mistral it over-reads the true precision-only disagreement by a factor of \textasciitilde{}1.9 while in Llama it under-reads by a factor of \textasciitilde{}0.62. See Section 4.3 for the full development.}
\label{fig:5}
\end{figure}

\subsubsection{Layer 1: the coarse method effect is real but model-dependent}

The cleanest comparison of comprehension vs generation as extraction methods --- both at fp16, both with protocol-matched generation parameters --- gives $\rho$ = 0.091 in Mistral 7B and $\rho$ = 0.355 in Llama 3.1 8B. In Mistral, the comprehension and generation pipelines produce nearly orthogonal RDMs (a Spearman correlation of 0.09 is below typical chance variation in our analysis); in Llama 3.1, they produce moderately correlated but still very different RDMs. Jeong (2026a)'s original reading --- that comprehension and generation are different things --- survives at the qualitative level in both models. But it survives much more strongly in Mistral than in Llama 3.1. Whatever "method effect" means here, it is not a single number; it is a model-dependent property.

\subsubsection{Layer 2: sub-parameter sensitivity is robust and large}

When we hold both method (generation) and precision (fp16) constant and only change the eight protocol parameters that Jeong (2026a)'s pipeline used (five-template prompts vs single prompt, ten samples vs one, greedy vs stochastic decoding, max\_new\_tokens 256 vs 100, mid-generation vs last-token extraction, middle layer vs best layer, chat-template-on vs chat-template-off, neutral-baseline centering vs global-mean centering), the resulting RDMs correlate at $\rho$ = 0.019 in Mistral 7B and $\rho$ = 0.033 in Llama 3.1 8B. Both numbers are at floor: changing implementation choices that one might naively assume to be "engineering details" produces RDMs that are essentially uncorrelated with the RDMs produced under different choices, and this is true in both models we have data for. This is the most robust of the four layers, in the sense that it gives the same qualitative answer across both models, and it is also the largest in magnitude (smallest $\rho$): sub-parameter choices within generation extraction account for more of the disagreement than the method label itself does.

This is a result that, on its own, should give pause to anyone reading the existing emotion-vector literature. Two studies that both describe their method as "generation-based extraction" can produce essentially uncorrelated emotion RDMs depending on which subset of these eight parameters they happened to use, even if precision and architecture are identical. The current practice of describing the method at a high level ("we extract emotion vectors from generated samples") is not specific enough to make results comparable across papers.

\subsubsection{Layer 3: the true precision effect is real and model-dependent}

When we hold method (generation) and all eight sub-parameters constant and only change precision from fp16 to INT8 --- the cleanest possible isolation of the quantization effect --- the resulting RDMs correlate at $\rho$ = 0.215 in Mistral 7B and $\rho$ = 0.527 in Llama 3.1 8B. Both are well above the sub-parameter floor of layer 2 (so changing precision while holding everything else constant is a smaller perturbation than changing implementation details within fp16 generation), but neither is close to the kind of within-protocol agreement that would license treating fp16 and INT8 results as interchangeable. The Mistral number (0.215) indicates that quantization fundamentally rearranges the cosine RDM in a 7B model --- most of the structure is gone --- while the Llama 3.1 number (0.527) indicates substantial but not destructive change in an 8B model. The two-fold gap between the two models is important: it says that quantization effects on emotion geometry are not a single number that can be characterized once and applied to all models. They depend on the model.

This is the layer that most directly motivates a dedicated quantization study, and we treat the present result as a pilot rather than as a full account. Two models with two precision conditions each is enough to establish that the effect is real and that it is model-dependent; it is not enough to map out how the effect scales with model size, with architecture family, with quantization scheme (NF4 vs INT8 vs others), or with which layer of the model the extraction is performed at.

\subsubsection{Layer 4: the conflated statistic lies in opposite directions for the two models}

The comparison that Jeong (2026a) actually reported for its 7B+ models is the one between A (comprehension at fp16) and D (generation at INT8) --- comparing the new comprehension pipeline to its own generation pipeline directly, without isolating which of the three intervening differences (method, sub-parameters, precision) account for what. This is the \textit{total} cross-experiment statistic, and it gives $\rho$ = 0.409 in Mistral 7B and $\rho$ = 0.325 in Llama 3.1 8B.

The structure of the result is the methodological lead of this paper. Compare each conflated number to its corresponding "true precision effect" from layer 3:

\begin{itemize}
\item \textbf{Mistral 7B}: conflated A vs D = 0.409, true C vs D = 0.215. A reader who interpreted the conflated A-vs-D statistic as a measure of the precision effect would over-estimate the true precision-only disagreement by a factor of \textasciitilde{}1.9, because the comprehension-vs-generation method effect (layer 1) and the sub-parameter effect (layer 2) both contribute additional disagreement on top of precision alone. The A-vs-D statistic is not an estimator of precision; it is a mixture, and treating it as a precision estimator produces the over-reading.
\end{itemize}

\begin{itemize}
\item \textbf{Llama 3.1 8B}: conflated A vs D = 0.325, true C vs D = 0.527. Here a reader who interpreted the conflated statistic as a precision measure would \textit{under-estimate} the true precision-only disagreement by a factor of \textasciitilde{}0.62, because in this model the precision effect on its own is moderate but the method and sub-parameter effects pull the cross-experiment $\rho$ further down. Again, the A-vs-D statistic is not an estimator of precision; it is the same mixture as in Mistral, but the relative magnitudes of the three contributing layers happen to align in a way that makes the mixture \textit{smaller} than precision alone, not larger.
\end{itemize}

The conflated cross-experiment statistic therefore distorts in \textbf{opposite directions for the two models} when read as a precision measure: it would be over-read as a precision measure in Mistral and under-read as a precision measure in Llama 3.1. A reader of the prior literature who saw only the conflated number for Mistral would conclude that quantization had a moderately bad effect on the geometry; a reader who later saw a similar conflated number for Llama 3.1 would conclude that the effect was about the same magnitude. Both readings would be wrong, in different directions, and the only way to recover the truth is the four-layer decomposition we report here.

\subsubsection{The general lesson}

Layer 4 is the practical lesson of this section. A cross-experiment $\rho$ between two prior emotion-vector studies --- even if both are well-documented and use overlapping models --- does not tell you what a single underlying difference does, because such a number always conflates multiple differences and the conflations distort in opposite directions for different models. The only way to interpret a cross-experiment $\rho$ honestly is to break it into its layered constituents: hold everything constant and change one thing at a time.

This is also the practical reason that Jeong (2026a) does not need to be treated as an error. Its central observation was that comprehension and generation pipelines disagree, and that observation survives in our layer 1 result for Mistral ($\rho$ = 0.091) and is partially echoed in Llama 3.1 ($\rho$ = 0.355). Where Jeong (2026a) stopped, however, was at the conflated cross-experiment statistic; the present paper continues by adding the protocol-matched fp16 generation extraction and showing that the method effect, the sub-parameter effect, and the precision effect are three distinct things rather than one. The right reading of Jeong (2026a) in light of these results is that it correctly identified the qualitative phenomenon and that the present paper supplies the structure needed to interpret the magnitude. We emphasize that this is an extension, not a correction.

We want to be precise about the scope of this reinterpretation. The precision confound we document is specifically scoped to the 7B+ subset of Jeong (2026a)'s dataset --- that is, Mistral 7B and Llama 3.1 8B, the two models that paper extracted at INT8 under a VRAM constraint. Its smaller models (Llama 3.2 3B, SmolLM2 1.7B, and the sub-2B models) were extracted at fp16 in the first place, so the fp16-vs-INT8 layer of our decomposition does not apply to them. The method-effect layer and the sub-parameter-sensitivity layer would in principle still apply to those smaller models if tested (we do not test them), but the precision layer --- the one on which layer 4's conflated-statistic distortion depends --- is a phenomenon of the 7B+ subset only. Any reinterpretation of Jeong (2026a) in light of the present findings should respect this boundary: the precision-confound lesson is 7B+ specific, while the method- and sub-parameter-effect lessons are more general.

We also note three limitations of the four-layer analysis. First, it is based on n = 2 models, which is enough to establish that the structure is model-dependent but not enough to characterize how that dependence works. Second, the precision contrast is fp16 vs INT8 specifically, and we have no data on intermediate precisions (4-bit quantization schemes, mixed-precision schemes) or on the question of whether the effect scales linearly with bit depth or saturates. Third, the protocol-matched fork pins eight specific sub-parameters and treats them as a single bundle; we have not isolated which of the eight contributes most to the layer 2 effect, and a more granular ablation would be a natural follow-up.

\begin{figure}[H]
\centering
\setcounter{figure}{5}
\includegraphics[width=\linewidth]{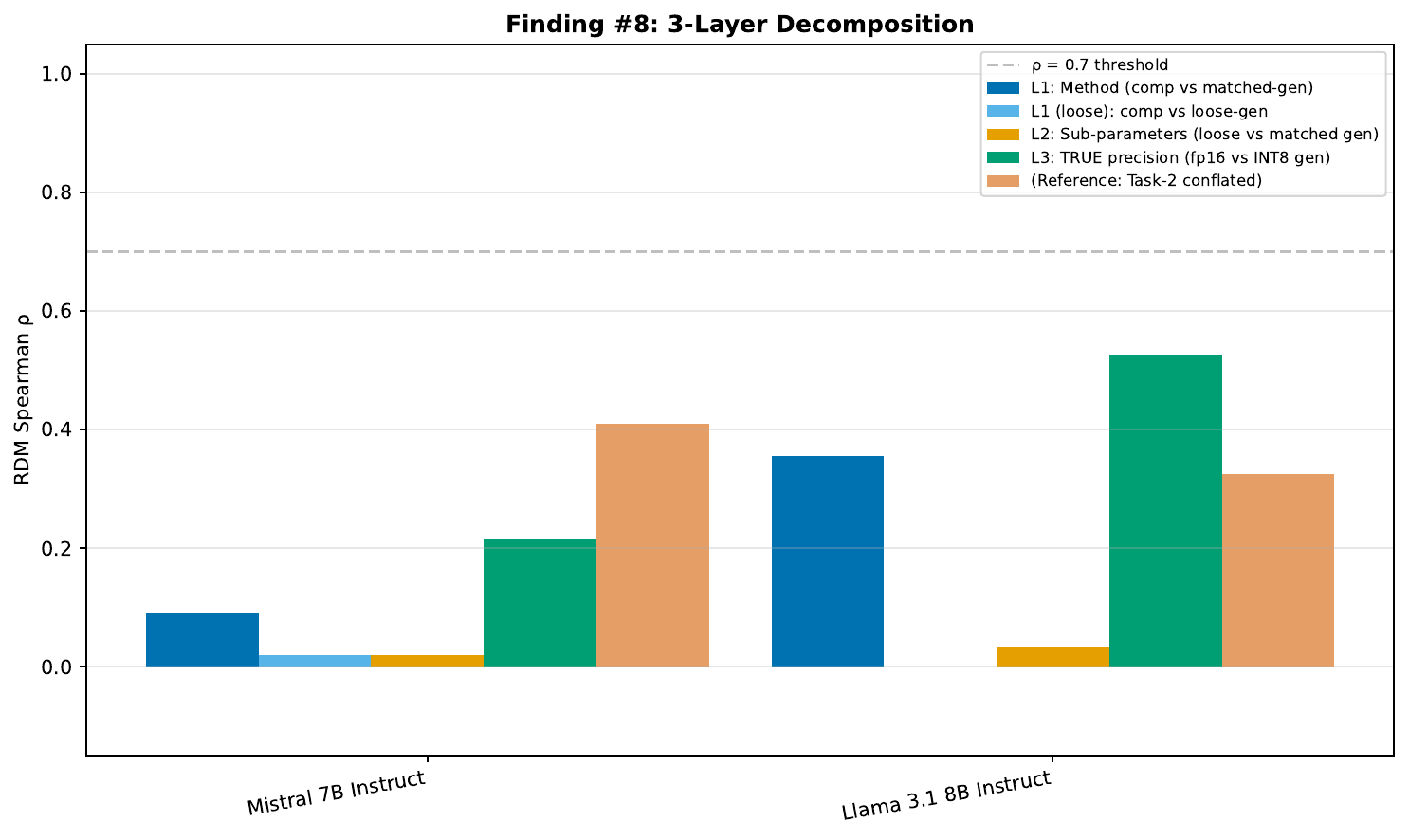}
\caption{\textit{Four-layer decomposition as grouped bar chart.} Per-model Spearman $\rho$ for the four isolating contrasts of Figure 5 and Section 4.3, shown as grouped bars for direct layer-by-layer comparison between Mistral 7B Instruct and Llama 3.1 8B Instruct. Layers 1 (method, A vs C), 2 (sub-parameters, B vs C), and 3 (precision, C vs D) are the three clean isolating contrasts; layer 4 (conflated A vs D) is shown in a distinct color to indicate that it is a mixture rather than an estimator. The layer-2 (sub-parameter) bars are the lowest for both models and nearly identical ($\rho$ = 0.019 vs 0.033), establishing sub-parameter sensitivity as the largest robust source of disagreement within generation extraction. The layer-3 (precision) bars show a factor-of-\textasciitilde{}2.5 gap between the two models (0.215 vs 0.527), establishing that quantization effects on emotion geometry are model-dependent. The layer-4 (conflated) bars distort in opposite directions relative to layer 3 --- over-reading in Mistral and under-reading in Llama --- which is the practical danger that motivates the four-layer decomposition. See Section 4.3 and Figure 5 for the full matrix.}
\label{fig:6}
\end{figure}

These three methodological findings together provide the infrastructure needed to read the substantive results of Section 3 safely. The anisotropy reliability flag of Section 4.1 explains why Gemma-3 1B base is excluded from the universality claim and treated as a maturity-gradient endpoint instead. The cross-backend equivalence of Section 4.2 explains why our four HF-backend models can be freely compared with our eight TL-backend models throughout Section 3. The four-layer decomposition of Section 4.3 explains why we hold extraction pipeline, precision, sub-parameters, and analysis protocol identical across all twelve models, and why a less disciplined cross-model comparison would have produced numbers that could not be interpreted at all. Section 5 turns to the broader implications of the substantive findings, the relationship of this paper to Jeong (2026a), and the limitations of our dataset.

\section{Discussion}

This paper has presented a foundational cross-model study of 21-emotion vector geometry in twelve small language models, holding extraction pipeline, precision, and analysis protocol constant across the dataset. The substantive picture (Section 3) is that 21-emotion geometry is shared across five mature SLM architectures at high RDM correlations ($\rho$ = 0.74--0.92), is robust to large differences in measured behavior (Qwen 2.5 vs Llama 3.2 on opposed MTI Compliance facets, $\rho$ = 0.81), sits on top of a continuous representation maturity gradient strongly correlated with model size, and is preserved at the family level across a 2.7$\times$ scale gap (Llama 3.2 $\times$ Llama 3.1, $\rho$ = 0.92). The methodological picture (Section 4) is that this characterization can only be done safely under three controls: an anisotropy reliability flag rather than a normalization correction, an explicit cross-backend equivalence check when the dataset mixes extraction tools, and a four-layer decomposition of what reads from a distance as a single "method effect" between comprehension and generation extraction. This section discusses what the combined picture means, how it relates to prior work, what it leaves unanswered, and where its limits are.

\subsection{What the shared geometry tells us, and what it does not}

The central empirical claim of this paper is that five SLMs from architecturally distinct families, spanning 1.5B to 8B parameters, encode 21 emotion concepts at near-identical cosine geometries when measured under a unified comprehension extraction pipeline. The strength of this claim depends on what "near-identical" means here, and we want to be precise. It does not mean that the five models have the same emotion vectors in any absolute sense --- they live in residual streams of different dimensionality, with different basis directions, and a vector from one model cannot be moved to another without a learned alignment. What it means is that the \textit{relational} structure of the 21 emotions, expressed as the 21 $\times$ 21 cosine RDM, is preserved across the five models at correlations that are at the top of what one would expect even from repeated runs of related procedures. The shape of the emotion space is the same shape, even when the coordinate systems differ.

This is informative for several reasons. For interpretability research, it suggests that techniques developed for analyzing the affective representations of one model have a defensible basis for transfer to other models in the same regime, which is not something one could assume \textit{a priori}. For safety work, it means that emotion-targeted interventions characterized on one SLM are likely to operate on a representation of similar shape in another SLM, modulo the alignment problem above. For our broader research program, it provides a representational baseline against which putative "model-exclusive" affective features can be defined: features that exist in one model and not in others, on a substrate that is otherwise shared, are exactly the features that mechanistic comparison should be looking for, and a foundational paper of this kind is what makes that comparison well-defined.

What the shared geometry does not tell us is \textit{why} it is shared. Three explanations are consistent with the data and we cannot adjudicate between them on the present results alone. The first is that the language modeling objective itself, when applied to text corpora that contain human emotional content, tends to discover roughly the same arrangement of these concepts regardless of architectural details. The second is that the training corpora used by current open-weight SLMs overlap heavily enough that the emotion structure is being inherited from data rather than discovered independently. The third is that the geometry reflects properties of the human language being modeled --- the way English (and the smaller fraction of other languages in these corpora) encodes affect --- rather than properties of either the training objective or the architecture. The three explanations make different predictions for models trained on different corpora, for non-English language modeling, and for synthetic data, and our dataset speaks to none of them. We treat the cross-architecture universality result as a phenomenon to be explained, not as one we have explained.

\subsection{Behavioral dissociation and the location of "compliance"}

The behavioral--representational dissociation between Qwen 2.5 1.5B Instruct and Llama 3.2 3B Instruct (Section 3.2) is, on its face, a single data point: two models, opposite poles on two MTI Compliance facets, $\rho$ = 0.81 between their emotion RDMs. The pair was selected precisely because it is the strongest within-dataset behavioral contrast we have access to, and we acknowledged in Section 3.2 that this is a stress test rather than a population study. The finding that even \textit{this} contrast does not appear in emotion geometry is therefore a constraint on theories of model behavior, not a complete account.

The constraint it imposes is the following. Any account of compliance behavior that would predict it to be tightly coupled to a model's low-level affective representation has to contend with the fact that two models with diametrically opposed compliance profiles are using nearly the same affective representation. The differences must arise \textit{somewhere}, and the residual stream at the best layer for emotion vector extraction is not where they arise. Plausible candidate locations include the policy layer that decides what to do with the representation's contents, the role-play and persona conditioning that is reshaped by RLHF, the token-level decoding dynamics that we are not measuring here, and any number of attention-mediated downstream computations. We do not have a mechanistic account of which of these is the actual locus, and we do not claim one. The claim is the negative one: it is not the emotion representation itself.

This negative claim has a positive corollary that we find more interesting than the negative form. If behavioral compliance is not coupled to emotion geometry but two SLMs nonetheless use very similar emotion geometries, then the \textit{route} from a similar emotion representation to two opposite behavioral profiles is itself a research target. Two models being given the same representational input and producing opposite outputs is the cleanest possible setup for a mechanistic interpretability study of where the divergence comes from. The MTI scoring infrastructure makes the behavioral end of this comparison concrete; the present paper makes the representational end concrete; what is missing is the work between the two ends.

\subsection{The maturity gradient and what RLHF actually does}

The Gemma-3 1B contrast (Section 3.3) and the size correlations (Section 3.5) together describe a representation maturity gradient that scales with model size in the sharpness dimension (anisotropy and RDM std fall sharply with \texttt{d\_\allowbreak{}model} and parameter count) but not in the location dimension (best-layer percentage is family-determined and only weakly correlated with size). On this gradient, Gemma-3 1B base sits at the lower endpoint with anisotropy 0.997 and RDM std 0.333; the three mature small-to-mid families occupy the middle; the two 7B+ models sit at the upper end. RLHF acts substantially on Gemma-3 1B, shifting all four geometric descriptors and moving the steering regime from explosive to surgical, while leaving the five already-mature families essentially unchanged.

The simplest reading of these two facts together is that RLHF is, at least in the geometric sense we are measuring, a \textit{structuring} operation that has substantial effect only on representations that are not yet structured. Mature models do not need RLHF to have a usable emotion geometry, and the post-training they receive does not reshape that geometry meaningfully. Immature models have an emotion structure that is essentially absent at the geometric level, and post-training installs one. This is consistent with --- and provides a different angle of evidence on --- the \textit{Superficial Alignment Hypothesis} introduced by Zhou et al. (2023) and developed further by Lin et al. (2024), which argues from data-efficiency results (LIMA: 1,000 examples suffice for high-quality alignment) and from token-distribution analysis (alignment-tuned and base models decode nearly identically except on stylistic tokens) that alignment tuning primarily exposes capabilities already present in the pretrained base rather than installing new ones. Our finding adds a complementary piece to this picture at the level of one specific kind of representation: when the base representation is already structured, RLHF leaves it essentially unchanged at the geometric level; when the base representation is unstructured (the Gemma-3 1B case), RLHF substantially restructures it. The two regimes give different geometric pictures of what post-training actually does, and the cross-regime contrast is what makes the differential effect visible.

The reading is fragile in one important way: it rests on n = 1 in the immature regime. Gemma-3 1B is the only model in our dataset whose base form sits unambiguously above the 0.95 anisotropy threshold, and a single immature case cannot establish that the maturity gradient generalizes. What it can establish is that the contrast exists and has the qualitative character we describe; what it cannot establish is whether other immature SLMs (e.g. other very small base models) would show similar before-and-after patterns under RLHF. We treat the differential RLHF effect as a hypothesis the present paper raises and as one that a follow-up with more immature-regime models could test.

A second open question concerns the location dimension. Best-layer percentage is family-determined in our dataset (Llama-family at \textasciitilde{}38\%, Qwen at \textasciitilde{}54\%, Mistral at \textasciitilde{}30--40\%, SmolLM2 at \textasciitilde{}50\%, Gemma instruct at \textasciitilde{}58\%) and only weakly tracks size. This is an interesting fact in its own right: it says that \textit{where} in the network a family chooses to encode the affective structure is a property the family carries, and one that does not change much across base/instruct or across scale within the family. We do not have an account of why families differ on this dimension. Plausible explanations include differences in tokenizer (which changes how much work the early layers have to do before semantically informative representations emerge), in attention pattern, in normalization placement, and in the depth at which late-layer task heads start to pull representations toward output-token statistics. None of these is testable on the present dataset alone.

\subsection{Relationship to Jeong (2026a)}

Jeong (2026a) --- a nine-model SLM methodological comparison across five architectural families --- reported a strong dissociation between comprehension- and generation-based emotion vector extraction and treated it as primarily a method effect. That paper extracted its two 7B+ models (Mistral 7B and Llama 3.1 8B) at INT8 precision under a VRAM constraint and its smaller models at fp16. The present paper extends Jeong (2026a) in three ways. First, we show that the qualitative method effect survives at fp16 in Mistral 7B (Section 4.3, layer 1: $\rho$ = 0.091), so the original observation is not an artifact of quantization. Second, we show that what looks like a single "method effect" decomposes into three distinct sources of variance --- the method label itself, sub-parameter choices within generation, and precision --- which have different magnitudes and different model-dependence patterns. Third, we show that the conflated cross-experiment statistic that pools these three sources distorts in opposite directions for different models when read as a precision measure (Mistral over-read, Llama 3.1 under-read relative to the true precision-only effect), which means that any interpretation of a single $\rho$ value between two prior emotion-vector studies is unsafe without the layered decomposition.

We treat this as an extension of Jeong (2026a) rather than as a correction. The original observation --- that comprehension and generation pipelines disagree --- survives in our cleaner contrasts. What was missing was the structure underneath that observation, and the structure required exactly the kind of multi-model, multi-precision data that this paper provides. Jeong (2026a)'s INT8 choice for its 7B+ models was a reasonable response to a VRAM constraint at the time and is not the source of the dissociation it reported; the dissociation is real and is partly a method effect, partly a sub-parameter effect, and partly a precision effect, in proportions that depend on the model. Section 4.3.5 develops the scope boundary in detail: the precision layer of the three-way decomposition is specific to the 7B+ subset of Jeong (2026a)'s dataset, while the method and sub-parameter layers are more general. This is the right reading of Jeong (2026a) in light of the present results, and it is the framing we adopt throughout the paper.

One specific aspect of Jeong (2026a) that the present decomposition reframes is its headline result that generation-based extraction produces statistically superior emotion separation (Mann-Whitney \textit{p} = 0.007, Cohen's \textit{d} = $-$107.5 in that paper). That result was measured under a specific choice of the eight sub-parameters we enumerate in Section 2.3, and our layer 2 analysis shows that \textit{which} sub-parameter configuration one uses within the generation family drives cross-configuration $\rho$ to floor (0.02--0.03 in both Mistral 7B and Llama 3.1 8B). The superiority finding of Jeong (2026a) should therefore be read as specific to its sub-parameter configuration, not as a general property of generation-based extraction across the configuration space. A full mapping of how emotion-separation quality varies across the sub-parameter space would be the natural follow-up, and would clarify whether generation is superior across the space, at a single point in it, or in some identifiable sub-region. The present paper does not settle this question; it only establishes that the question is real and that the existing answer is conditional on a specific configuration choice.

We also note what the present paper does \textit{not} settle about Jeong (2026a). The three-source decomposition is based on n = 2 models and only on the specific INT8 vs fp16 precision contrast. Whether the same three-source structure holds at other quantization schemes, whether the model-dependence pattern generalizes beyond Mistral 7B and Llama 3.1 8B, and whether the eight sub-parameters can be decomposed into a smaller set of dominant contributors are all questions that the present data is not large enough to answer. A more granular ablation of the eight sub-parameters and a broader set of precision conditions would be a natural extension of this layer of the analysis.

\subsection{Limitations}

We see five main limitations of this work.

First, \textbf{n = 6 architectural families}, of which only five are represented in the universality result of Section 3.1. Five mature families is enough to falsify the strong claim that emotion geometry is architecture-specific --- it is not, at least within the open-weight SLM regime --- but it is not enough to characterize what aspects of architecture matter. A larger dataset that includes families we did not test (e.g. Phi, Yi, DeepSeek, Falcon, OLMo, or non-decoder architectures) would either strengthen the universality result or identify the architectural axis on which it breaks.

Second, \textbf{a single member of the Gemma family}. The maturity gradient claim of Section 3.3 and the differential RLHF effect rest on Gemma-3 1B specifically. We do not know whether the immature-regime characterization generalizes to other very small base models, and we cannot distinguish "Gemma-3 1B has an unusually under-structured base representation" from "all SLMs at or below 1B parameters have under-structured base representations." Adding other 1B-class base models, whether from other Gemma generations or from other families, would settle this. This single case also carries the fp16 precision artifact caveat noted in Sections 2.5 and 3.3: a bf16 or fp32 re-extraction of Gemma-3 1B base would be needed to disentangle representational from numerical contributions to its extreme anisotropy and RDM standard deviation.

Third, \textbf{n = 2 models for the layer 3 precision contrast}. The precision analysis of Section 4.3 (layer 3) relies on only two models for the cleanest contrast (protocol-matched fp16 vs INT8 generation). Two models is enough to establish that the precision effect is real and that it is model-dependent --- both numbers are well above the sub-parameter floor and they differ from each other by a factor of \textasciitilde{}2.5 --- but it is not enough to characterize how that dependence works. A precision study with more models, more quantization schemes, and a more granular sub-parameter ablation would fill this in.

Fourth, \textbf{the behavioral test in Section 3.2 is a single pair}. We chose the Qwen vs Llama 3.2 pair because it is the strongest within-dataset opposition we have on MTI Compliance facets, but a single pair cannot establish that no behavioral dimension whatsoever appears in emotion geometry. It can only establish that the strongest contrast we have access to does not appear. A broader behavioral--representational study using more model pairs and the full MTI axis structure would be needed to map out where, if anywhere, behavioral facets do leave a representational signature.

Fifth, \textbf{comprehension-mode extraction as the primary pipeline}. The cross-model results of Section 3 use comprehension-mode extraction throughout, on the grounds that it avoids confounding emotion representation with first-person generation conditioning. Section 4.3 shows that generation-mode extraction would have given a different picture --- substantially so for some models --- and we have not characterized which of the two pipelines is "more correct" in any absolute sense. Both are measurements of the same underlying residual stream at slightly different conditions, and the choice of comprehension as primary is justified for cross-model comparability but does not entitle us to claim that the generation-mode geometry is wrong. A version of the present paper that ran both pipelines on all twelve models in parallel would let us characterize the comprehension--generation gap as a function of model and architecture, which we have only done for Mistral 7B and Llama 3.1 8B in the protocol-matched fork.

Beyond these five, two further limitations are worth noting briefly. The 21-emotion vocabulary is fixed throughout and --- apart from the added \textit{neutral} category --- is the same vocabulary used in Jeong (2026a); whether the universality result extends to a substantially different affective vocabulary (e.g., one that includes self-conscious emotions, social emotions, or affect terms drawn from non-Western psychological traditions) is an open question. And the entire analysis is conducted in English; the cross-model picture for affective representations in models trained on substantially non-English corpora is unknown.

The present paper does one thing: it characterizes the shared 21-emotion geometry across twelve open-weight SLMs under a unified extraction protocol, and it documents the methodological infrastructure required to make that characterization safely. The picture it produces --- universality across mature architectures, robustness to opposed behavioral profiles, a maturity gradient that scales with size, family identity preservation, and a four-layer structure underneath what looked like a single methodological confound --- is, we believe, the foundational baseline that further cross-model SLM emotion-vector work needs to start from. The questions it leaves open are at least as interesting as the ones it answers, and we hope the dataset, the analysis pipeline, and the methodological controls released alongside the paper are useful infrastructure for that further work.

\appendix
\newpage
\begin{center}
\vspace*{2em}
{\Large \textbf{Appendices}}
\vspace{0.5em}
\hrule
\vspace{1.5em}
\end{center}
\section{Dataset Disclosure and Per-Model Metadata}

\subsection{Exact Model Identifiers and Provenance}

All twelve models are loaded by HuggingFace identifier at fp16 precision. Table A1 lists the exact identifier, architectural family, parameter count, layer count, hidden dimension, extraction backend, and the four geometric descriptors at best layer reported in the main text. The extraction was performed between 2026-04-08 and 2026-04-09 on Apple Silicon (local) and RunPod A40 (7B+ models); see Section 2.3 for pipeline details.

\textbf{Table A1. The twelve-model dataset.} B = base (pre-trained only); I = instruct (post-trained via RLHF or SFT+DPO variants, per each family's release). Best layer is given both as an absolute layer index and as a percentage of total depth. Anisotropy is measured on twenty short neutral declarative sentences at the best layer. RDM standard deviation is the standard deviation of off-diagonal entries of the 21 $\times$ 21 cosine RDM at the best layer. Backend: TL = TransformerLens, HF = raw HuggingFace transformers hooks (Section 2.4).

\begin{table}[H]
\centering
\footnotesize
\resizebox{\linewidth}{!}{%
\begin{tabular}{llllllllllll}
\toprule
\# & HuggingFace \texttt{model\_\allowbreak{}id} & Family & Variant & Size (B) & Layers & \texttt{d\_\allowbreak{}model} & Backend & Best Layer & Best Layer \% & Anisotropy & RDM std \\
\midrule
1 & \texttt{Qwen/\allowbreak{}Qwen2.5-1.5B} & Qwen 2.5 & B & 1.5 & 28 & 1536 & TL & 15 & 53.6 & 0.830 & 0.127 \\
2 & \texttt{Qwen/\allowbreak{}Qwen2.5-1.5B-Instruct} & Qwen 2.5 & I & 1.5 & 28 & 1536 & TL & 15 & 53.6 & 0.848 & 0.130 \\
3 & \texttt{HuggingFaceTB/\allowbreak{}SmolLM2-1.7B} & SmolLM2 & B & 1.7 & 24 & 2048 & HF & 13 & 54.2 & 0.854 & 0.123 \\
4 & \texttt{HuggingFaceTB/\allowbreak{}SmolLM2-1.7B-Instruct} & SmolLM2 & I & 1.7 & 24 & 2048 & HF & 11 & 45.8 & 0.808 & 0.118 \\
5 & \texttt{google/\allowbreak{}gemma-3-1b-pt} & Gemma-3 & B & 1.0 & 26 & 1152 & TL & 9 & 34.6 & 0.997 & 0.333 \\
6 & \texttt{google/\allowbreak{}gemma-3-1b-it} & Gemma-3 & I & 1.0 & 26 & 1152 & TL & 15 & 57.7 & 0.982 & 0.165 \\
7 & \texttt{meta-llama/\allowbreak{}Llama-3.2-3B} & Llama 3.2 & B & 3.0 & 28 & 3072 & TL & 11 & 39.3 & 0.649 & 0.104 \\
8 & \texttt{meta-llama/\allowbreak{}Llama-3.2-3B-Instruct} & Llama 3.2 & I & 3.0 & 28 & 3072 & TL & 11 & 39.3 & 0.627 & 0.103 \\
9 & \texttt{mistralai/\allowbreak{}Mistral-7B-v0.3} & Mistral 7B & B & 7.0 & 32 & 4096 & HF & 13 & 40.6 & 0.532 & 0.097 \\
10 & \texttt{mistralai/\allowbreak{}Mistral-7B-Instruct-v0.3} & Mistral 7B & I & 7.0 & 32 & 4096 & HF & 13 & 40.6 & 0.491 & 0.093 \\
11 & \texttt{meta-llama/\allowbreak{}Llama-3.1-8B} & Llama 3.1 & B & 8.0 & 32 & 4096 & TL & 12 & 37.5 & 0.642 & 0.096 \\
12 & \texttt{meta-llama/\allowbreak{}Llama-3.1-8B-Instruct} & Llama 3.1 & I & 8.0 & 32 & 4096 & TL & 12 & 37.5 & 0.680 & 0.106 \\
\bottomrule
\end{tabular}
}
\end{table}

\subsection{Backend Assignment Rationale}

Eight of the twelve models (Qwen 2.5 base/instruct, Gemma-3 base/instruct, Llama 3.2 base/instruct, Llama 3.1 base/instruct) were extracted via TransformerLens (TL), which provides named hook points on the residual stream and is the preferred backend wherever supported. The remaining four models (SmolLM2 1.7B base/instruct and Mistral 7B v0.3 base/instruct) were extracted via raw HuggingFace transformers hooks because TransformerLens does not currently support either architecture. The decision to fall back to HF raw hooks for these four models rather than drop them from the dataset preserves the architectural diversity of the mature-family set; the cross-backend equivalence test of Section 2.4 (cosine = 0.999998 on a matched Llama 3.2 3B Instruct unit test) licenses direct cross-backend comparison in Section 3.

\subsection{Best-Layer Stability Within Mature Families}

A useful summary observation visible in Table A1 is that the best layer, expressed as a percentage of total depth, is identical between base and instruct for four of the five mature families (Qwen 2.5: 53.6\% $\rightarrow$ 53.6\%; Llama 3.2: 39.3\% $\rightarrow$ 39.3\%; Mistral 7B: 40.6\% $\rightarrow$ 40.6\%; Llama 3.1: 37.5\% $\rightarrow$ 37.5\%). Only SmolLM2 shows a non-trivial shift (54.2\% $\rightarrow$ 45.8\%, equivalent to 2 absolute layer positions). By contrast, Gemma-3 1B shows a 23-percentage-point shift (34.6\% $\rightarrow$ 57.7\%) from base to instruct. This stability pattern complements the within-family RDM alignment pattern of Section 3.3 and further supports the interpretation that RLHF leaves mature emotion geometry essentially unchanged.

\subsection{Note on Mistral 7B Lineage}

An earlier extraction of \texttt{mistralai/\allowbreak{}Mistral-7B-v0.1} via TransformerLens was performed during pipeline development, when the TransformerLens supported-model list included v0.1 but not v0.3. Because the corresponding instruct model in the dataset is \texttt{mistralai/\allowbreak{}Mistral-7B-Instruct-v0.3}, this earlier extraction produced an asymmetric base/instruct pair from two different Mistral lineages (v0.1 base, v0.3 instruct) and is \textbf{not used in the main analyses of this paper}. All Mistral-related numbers reported in Sections 3, 4, and 5 are computed from the matched v0.3 base / v0.3 instruct pair extracted via raw HuggingFace hooks on 2026-04-09. The v0.1 base extraction is preserved in the supplementary data release as \texttt{mistralai\_\allowbreak{}Mistral-7B-v0.1/\allowbreak{}} for reproducibility of the pipeline development history, but should not be mixed with the v0.3 data for any downstream analysis.

\subsection{Numerical Stability Caveat for Gemma-3 1B Base}

As noted in Section 2.5 and Section 3.3, the Gemma-3 1B base model (\texttt{google/\allowbreak{}gemma-3-1b-pt}) exhibits per-layer mean cosine values that become NaN from approximately layer 12 onward under fp16 precision. The best layer (layer 9) sits within the numerically stable region, and the descriptors reported in row 5 of Table A1 are well-defined at that layer. However, anisotropy and RDM std at the 50\% and 75\% reference depths are not available for this single model, and we therefore exclude those two depths from any aggregate analysis that requires them. A bf16 or fp32 re-extraction of Gemma-3 1B base would resolve this caveat and is flagged as a robustness check in Section 5.5.

\subsection{Extraction Environment}

All twelve models were extracted in fp16 precision using a single unified pipeline. Hardware and software configuration is as follows.

\textbf{Cloud extraction (7B+ models).} Mistral 7B v0.3 base/instruct and Llama 3.1 8B base/instruct were extracted on RunPod NVIDIA A40 instances (48 GB VRAM, 50 GB container disk). The cloud environment used PyTorch 2.6.0+cu124 (forward-compatible with CUDA driver 12.4--12.8), TransformerLens 2.18.0, \texttt{transformers} 4.57.6 (pinned to <5 for TransformerLens compatibility), and Python 3.11. Mistral 7B v0.3 (both variants) used the raw HuggingFace \texttt{transformers} backend (\texttt{hf\_\allowbreak{}raw\_\allowbreak{}hooks} in the per-model metadata) because TransformerLens does not currently support v0.3; Llama 3.1 8B (both variants) used TransformerLens directly. Extraction script: \texttt{scripts/\allowbreak{}paper5\_\allowbreak{}extract.\allowbreak{}py}.

\textbf{Local extraction (smaller models).} Qwen 2.5 1.5B, SmolLM2 1.7B, Gemma-3 1B, and Llama 3.2 3B (base and instruct each) were extracted on Apple Silicon with the PyTorch MPS backend and the same software stack via the same \texttt{paper5\_\allowbreak{}extract.\allowbreak{}py} script, in fp16. Identical extraction protocol was applied throughout: comprehension via 21 emotion-labeled passages (Section 2.3), anisotropy baseline via 20 short neutral sentences (Section 2.5), and layer sweep with best-layer selection at the mean-cosine minimum.

\textbf{Protocol-matched generation fork.} The precision-isolation analysis of Section 4.3 used \texttt{scripts/\allowbreak{}paper5\_\allowbreak{}extract\_\allowbreak{}protocol\_\allowbreak{}matched.\allowbreak{}py}, a verbatim fork of Jeong (2026a)'s \texttt{emotion\_\allowbreak{}vector\_\allowbreak{}pilot\_\allowbreak{}v2.\allowbreak{}py} with a single substitution --- \texttt{dtype=torch.\allowbreak{}float16} replacing \texttt{quantize="int8"} --- so that fp16 vs INT8 can be contrasted while the eight sub-parameters of the generation pipeline are held identical. This fork was applied only to Mistral 7B Instruct and Llama 3.1 8B Instruct, the two 7B+ models in Jeong (2026a)'s dataset.

\textbf{Cross-platform consistency.} The v0.1 $\rightarrow$ v0.3 Mistral re-extraction documented in Appendix A.4 served as an implicit cross-platform reproducibility test: the same script, same dtype, and same protocol on RunPod produced size-correlation results (anisotropy vs \texttt{d\_\allowbreak{}model} $\rho$ = $-$0.882, RDM std vs size $\rho$ = $-$0.891) that are invariant to which Mistral version is used as the low-anisotropy endpoint, suggesting that neither the hardware switch (local Apple Silicon $\leftrightarrow$ cloud A40) nor the version switch (v0.1 $\leftrightarrow$ v0.3) introduces a substantive confound. Full per-model JSON metadata, including the software-version dictionary, is included in the supplementary code release.

\section*{References}
\setlength{\parindent}{0pt}
\setlength{\parskip}{0.5em}
\everypar{\hangindent=1.5em\hangafter=1}

Allal, L. B., Lozhkov, A., Bakouch, E., Bl\'{a}zquez, G. M., Penedo, G., Tunstall, L., Marafioti, A., Kydl\'{i}\v{c}ek, H., Lajar\'{i}n, A. P., Srivastav, V., Lochner, J., Fahlgren, C., Nguyen, X.-S., Fourrier, C., Burtenshaw, B., Larcher, H., Zhao, H., Zakka, C., Morlon, M., Raffel, C., von Werra, L., \& Wolf, T. (2025). \textit{SmolLM2: When Smol Goes Big --- Data-Centric Training of a Small Language Model.} arXiv:2502.02737.

Anthropic. (2026). \textit{Emotion Concepts and their Function in a Large Language Model.} Transformer Circuits Thread. https://transformer-circuits.pub/2026/emotions/

Dubey, A., Jauhri, A., Pandey, A., Kadian, A., Al-Dahle, A., Letman, A., et al. (2024). \textit{The Llama 3 Herd of Models.} arXiv:2407.21783.

Gemma Team. (2025). \textit{Gemma 3 Technical Report.} arXiv:2503.19786.

Jeong, J. (2026a). \textit{Extracting and Steering Emotion Representations in Small Language Models: A Methodological Comparison.} arXiv:2604.04064.

Jeong, J. (2026b). \textit{MTI: A Behavior-Based Temperament Profiling System for AI Agents.} arXiv:2604.02145.

Jiang, A. Q., Sablayrolles, A., Mensch, A., Bamford, C., Chaplot, D. S., de las Casas, D., Bressand, F., Lengyel, G., Lample, G., Saulnier, L., Lavaud, L. R., Lachaux, M.-A., Stock, P., Le Scao, T., Lavril, T., Wang, T., Lacroix, T., \& El Sayed, W. (2023). \textit{Mistral 7B.} arXiv:2310.06825.

Lin, B. Y., Ravichander, A., Lu, X., Dziri, N., Sclar, M., Chandu, K., Bhagavatula, C., \& Choi, Y. (2024). The Unlocking Spell on Base LLMs: Rethinking Alignment via In-Context Learning. In \textit{International Conference on Learning Representations (ICLR).} arXiv:2312.01552.

Mu, J., Bhat, S., \& Viswanath, P. (2018). All-but-the-Top: Simple and Effective Postprocessing for Word Representations. In \textit{International Conference on Learning Representations (ICLR).} arXiv:1702.01417.

Park, K., Choe, Y. J., \& Veitch, V. (2024). The Linear Representation Hypothesis and the Geometry of Large Language Models. In \textit{Proceedings of the 41st International Conference on Machine Learning (ICML)}, PMLR 235:39643--39666. arXiv:2311.03658.

Rimsky, N., Gabrieli, N., Schulz, J., Tong, M., Hubinger, E., \& Turner, A. M. (2024). Steering Llama 2 via Contrastive Activation Addition. In \textit{Proceedings of the 62nd Annual Meeting of the Association for Computational Linguistics (ACL)}, pages 15504--15522. arXiv:2312.06681.

Yang, A., Yang, B., Zhang, B., Hui, B., Zheng, B., Yu, B., Li, C., Liu, D., Huang, F., Wei, H., et al. (2025). \textit{Qwen2.5 Technical Report.} arXiv:2412.15115.

Zhou, C., Liu, P., Xu, P., Iyer, S., Sun, J., Mao, Y., Ma, X., Efrat, A., Yu, P., Yu, L., Zhang, S., Ghosh, G., Lewis, M., Zettlemoyer, L., \& Levy, O. (2023). LIMA: Less Is More for Alignment. In \textit{Advances in Neural Information Processing Systems (NeurIPS).} arXiv:2305.11206.

\end{document}